%% file: manuscript.tex
\documentclass[letterpaper]{article} % DO NOT CHANGE THIS
\usepackage{aaai2026}  % DO NOT CHANGE THIS
\usepackage{times}  % DO NOT CHANGE THIS
\usepackage{helvet}  % DO NOT CHANGE THIS
\usepackage{courier}  % DO NOT CHANGE THIS
\usepackage[hyphens]{url}  % DO NOT CHANGE THIS
\usepackage{graphicx} % DO NOT CHANGE THIS
\usepackage{natbib}  % DO NOT CHANGE THIS AND DO NOT ADD ANY OPTIONS TO IT
\usepackage{caption} % DO NOT CHANGE THIS AND DO NOT ADD ANY OPTIONS TO IT
\usepackage{algorithm}
\usepackage{algorithmic}
\usepackage{soul}
\usepackage[utf8]{inputenc}

\usepackage{multirow}
\usepackage{xcolor}
\usepackage{footnote}
\usepackage{graphicx}
\usepackage{amsmath}
\usepackage{amsthm}
\usepackage{booktabs}
\usepackage{algorithm}
\usepackage{algorithmic}
\usepackage[switch]{lineno}
\usepackage{amsfonts}
\usepackage{amsmath}
\DeclareMathOperator{\atantwo}{atan2}
\usepackage{newfloat}
\usepackage{listings}
\DeclareCaptionStyle{ruled}{labelfont=normalfont,labelsep=colon,strut=off} % DO NOT CHANGE THIS
\floatstyle{ruled}
\newfloat{listing}{tb}{lst}{}
\floatname{listing}{Listing}
\title{Rethinking Bias in Generative Data Augmentation for Medical AI: a Frequency Recalibration Method}
\author{
    Chi Liu\textsuperscript{\rm 1}\thanks{Corresponding Author (email: chiliu@cityu.edu.mo)},
    Jincheng Liu\textsuperscript{\rm 1},
    Congcong Zhu\textsuperscript{\rm 1},
    Minghao Wang\textsuperscript{\rm 1}, 
    Sheng Shen\textsuperscript{\rm 2}, 
    Jia Gu\textsuperscript{\rm 1}, 
    Tianqing Zhu\textsuperscript{\rm 1}, 
    Wanlei Zhou\textsuperscript{\rm 1}
}
\affiliations{
    \textsuperscript{\rm 1}Faculty of Data Science, City University of Macau, Macao SAR, China\\
    \textsuperscript{\rm 2}Design and Creative Technology vertical, Torrens University Australia, Ultimo, 2007, NSW, Australia\\
    \{chiliu,D21090103791,cczhu,mhwang,jiagu,tqzhu,wlzhou\}@cityu.edu.mo, sheng.shen@torrens.edu.au
}

\usepackage{bibentry}
\begin{document}

\maketitle

\begin{abstract}
Developing Medical AI relies on large datasets and easily suffers from data scarcity. Generative data augmentation (GDA) using AI generative models offers a solution to synthesize realistic medical images. However, the bias in GDA is often underestimated in medical domains, with concerns about the risk of introducing detrimental features generated by AI and harming downstream tasks. This paper identifies the frequency misalignment between real and synthesized images as one of the key factors underlying unreliable GDA and proposes the Frequency Recalibration (FreRec) method to reduce the frequency distributional discrepancy and thus improve GDA. FreRec involves (1) Statistical High-frequency Replacement (SHR) to roughly align high-frequency components and (2) Reconstructive High-frequency Mapping (RHM) to enhance image quality and reconstruct high-frequency details. Extensive experiments were conducted in various medical datasets, including brain MRIs, chest X-rays, and fundus images. The results show that FreRec significantly improves downstream medical image classification performance compared to uncalibrated AI-synthesized samples. FreRec is a standalone post-processing step that is compatible with any generative model and can integrate seamlessly with common medical GDA pipelines.
\end{abstract}

\section{Introduction}
Developing medical AI has become an essential practice for computer-aided disease diagnosis \cite{litjens2017survey,li2018automated}. This commonly involves a data-driven task that relies heavily on a large volume of medical images for model training. However, it often suffers from data scarcity due to privacy issues, high costs of collecting real-world medical images, and highly imbalanced class distributions \cite{rajpurkar2022ai,he2020deployment,wang2024economic}.
% Medical image classification is a fundamental task for computer-aided disease diagnosis and clinical analysis. A common practice is to develop advanced medical artificial intelligence (AI) models to automate the classification \cite{litjens2017survey}. This is a data-driven task that relies heavily on a large volume of medical images for model training. However, it often suffers from data scarcity due to privacy issues, high costs of collecting real-world medical images, and highly imbalanced class distributions \cite{rajpurkar2022ai}.

A promising solution to address the shortage of clinically available medical images is using generative AI models such as Generative Adversarial Networks (GANs) and Diffusion Models (DMs) to synthesize realistic images that augment existing training datasets \cite{dayarathna2024deep,wang2024towards}. Known as generative data augmentation (GDA), this technique increasingly produces large-scale synthetic medical images that closely resemble real data, thereby improving disease classification performance \cite{liu2020towards,gao2023synthetic,chen2024towards,shang2024synfundus1mhighqualitymillionscalesynthetic}. Moreover, GDA can generate diverse pathological samples from healthy data, enhancing cross-domain generalization, supporting few-shot learning, and mitigating classifier bias \cite{ktena2024generative}.

\begin{figure}
    \centering
    \includegraphics[width=\linewidth]{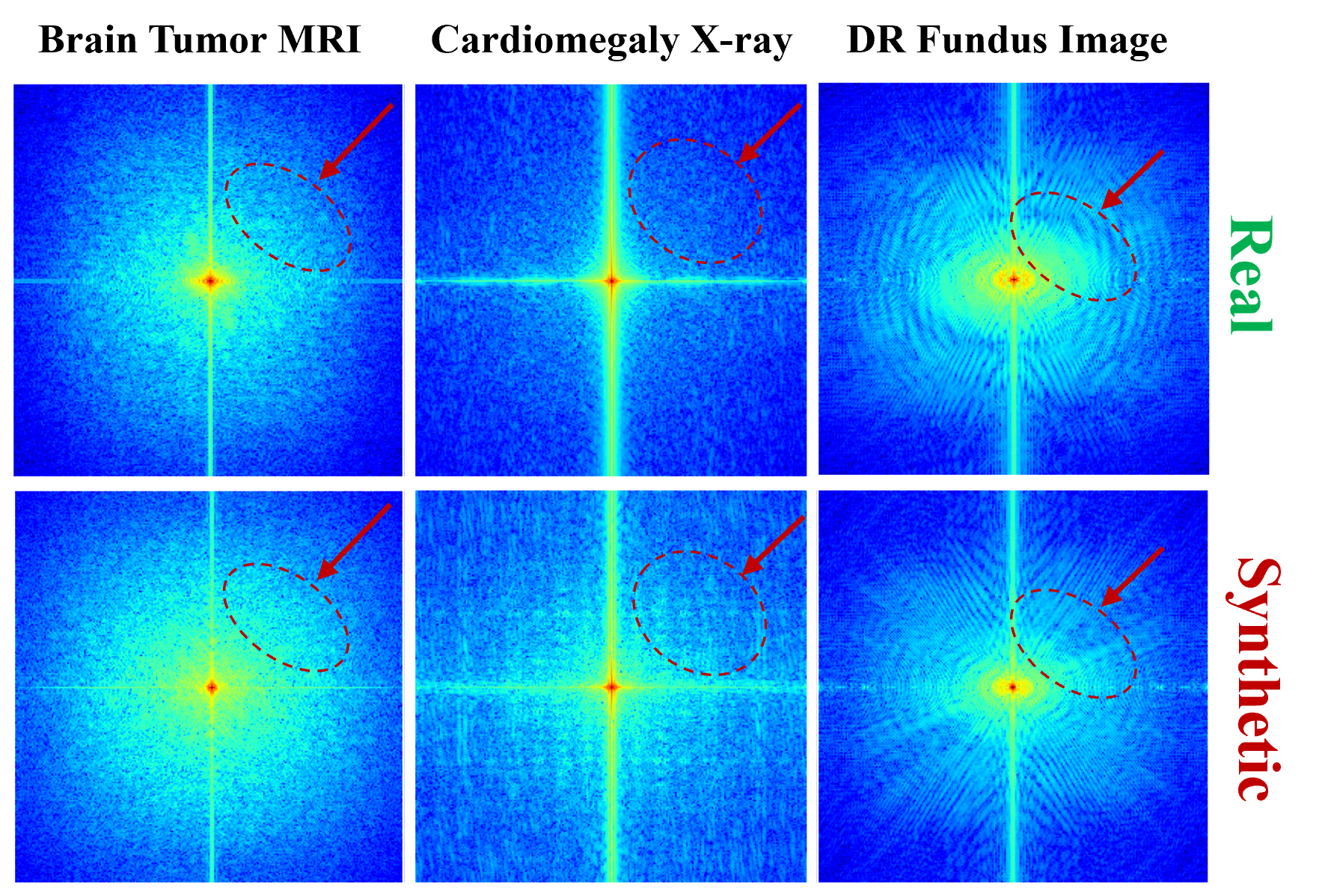}
    \caption{The average spectra of real and AI-synthetic images with different diseases and image types. The higher-frequency differences are discernible (e.g. as red circles indicate). Synthetic images are generated by different models: MRIs from FastGAN, fundus images from VC-Diffusion and X-rays from StyleGAN3.}
    \label{fig:spectrum}
\end{figure}

Despite being widely applied to developing medical AI, GDA faces growing concerns regarding its reliability. In general AI fields, AI-synthesized samples have proven to cause bias in certain tasks. For instance, language models can experience performance collapse after repeated training on generated content \cite{shumailov2024ai}. Likewise, computer vision models trained with GDA often fail to achieve consistent improvement through trial and error \cite{singh2024synthetic}. However, in the medical imaging domain, this phenomenon is generally overlooked. Alongside many medical AI studies having reported a positive effect of GDA, it remains unclear whether synthesized medical samples consistently benefit downstream tasks or whether they may introduce detrimental features generated by AI.

% Despite its achievements, GDA faces growing concerns regarding reliability. In general deep learning fields, AI-synthesized samples have shown questionable reliability in certain tasks. For instance, language models can experience performance collapse after repeated training on generated content \cite{shumailov2024ai}. Likewise, computer vision models trained with GDA often fail to achieve consistent improvement through trial and error \cite{singh2024synthetic}. In the medical imaging domain, however, the reliability of GDA is often underestimated despite widespread support for its potential. It remains unclear whether synthesized medical samples consistently benefit downstream tasks or if they introduce detrimental features generated by AI. 

Intrigued by this gap, in this paper, we provide a close look at the bias of GDA when applying in medical image classification tasks. Our empirical results show that GDA can sometimes be detrimental for downstream training, mirroring observations in general AI fields. Unlike a recent study of natural image tasks that addressed this bias as a domain shift without exploring its underlying cause \cite{10657264}, we investigate it from a frequency-domain perspective. Our analysis is inspired by recent research revealing high-frequency misalignment between real and AI-generated images~\cite{durall2020watch,dzanic2020fourier,9577744,corvi2023intriguing,frank2020leveraging}. Since medical images such as MRIs and X-rays are especially sensitive to high-frequency variations due to their imaging processes and reliance on subtle pathological details, they may be particularly vulnerable to such frequency gaps. 

Hence, we are motivated to posit this frequency gap (as shown in Fig. \ref{fig:spectrum}) as one of the key factors underlying unreliable GDA. Following this, we propose Frequency Recalibration (FreRec), a two-step method enabling coarse-to-fine alignment of frequency distributions between synthesized and real images. The first step, Statistic High-frequency Replacement (SHR), roughly aligns distributions by replacing the high-frequency components of AI samples with statistically sampled counterparts from real images. The second step, Reconstructive High-frequency Mapping (RHM), refines the perturbed samples from Step 1, reconstructing their high-frequency details by mapping onto the frequency distribution of natural images while enhancing the image quality for downstream use. Experiments on diverse medical image classification tasks demonstrate that FreRec-calibrated GDA samples significantly and consistently improve downstream performance compared to uncalibrated AI-synthesized samples. Notably, unlike previous methods that require retraining generative models with frequency-domain regularization to reduce their frequency bias \cite{durall2020watch, jiang2021focal}, FreRec is a standalone post-processing step compatible with any generative model, including GANs and diffusion models, making it a practical, plug-and-play, and cost-effective solution for medical AI pipelines.

% In a nutshell, the contributions of this paper include:
% \begin{itemize}
%     \item An in-depth analysis of unstable GDA, identifying that the frequency distributional discrepancy between real training samples and  AI-synthesized samples is a main cause.
%     \item A novel Frequency Recalibration method that can reduce the frequency distributional discrepancy and thus improve GDA's downstream reliability.  
%     \item Extensive experimental validations of the proposed method in diverse medical image classification datasets and tasks. 
% \end{itemize}

\section{Related work}
\subsection{Generative data augmentation for medical AI}
Using Generative AI models to synthesize medical data for dataset augmentation has become a common practice to address medical data scarcity in developing medical AI \cite{wang2024towards}. Successful applications can be identified in various medical domains, including X-ray scans \cite{liu2020towards,gao2023synthetic}, computed tomography (CT) \cite{chen2024towards}, fundus images \cite{shang2024synfundus1mhighqualitymillionscalesynthetic}, MRI \cite{dayarathna2024deep}, and even foundation models for multi-modality generation \cite{wang2024self}. The usage of GDA can also be extended to serve specific purposes such as privacy-preserving training \cite{guillaudeux2023patient} or fairness enhancement \cite{ktena2024generative}. However, although the reliability of AI-synthesized samples has recently been scrutinized in general deep learning fields \cite{shumailov2024ai, singh2024synthetic}, in medical domains, this risk remains underestimated and underexplored. A recent study addressed the issue from a domain-shift perspective but evaluated only natural images and did not investigate its root cause \cite{10657264}. 

\subsection{Frequency gap in AI-synthesized images}
Previous studies have identified a frequency gap between AI-generated and real images. Existing work \cite{xu2019frequency, wang2020high, rahaman2019spectral} has demonstrated that deep neural networks exhibit preference in learning information across different frequency bands, leading to frequency artifacts when applied to generating images. For example, Durall et al. \cite{durall2020watch} observed significant differences in the spectral distributions of real and GAN-generated images. Other research \cite{dzanic2020fourier,9577744,corvi2023intriguing, frank2020leveraging} has shown that this disparity is detectable in both phase and amplitude spectra and occurs in both GAN-generated and Diffusion-generated images. There are also studies proposing to address the generative frequency bias via a frequency-domain regularizer or a frequency loss \cite{durall2020watch, jiang2021focal}. But such methods require retraining the generative model, and thus are less practical for medical GDA considering cost-effectiveness and data availability.

\section{The Frequency Recalibration Method}
% \subsection{Overview}
This study is motivated by the intriguing generative frequency bias phenomenon, with a goal to answer the following questions: \textit{1) will the frequency gap between AI-generated and real images affect GDA in medical AI? 2) If so, is there a cost-effective, universal method to close the frequency gap to improve GDA for various medical image classification tasks?}

Since prior research shows that frequency abnormalities in AI-generated images often appear in higher frequency components, a pattern also observed in medical images (see Fig.\ref{fig:spectrum}), a potential improvement on GDA is post-processing AI-synthesized samples by recalibrating their high-frequency distribution to match real images. To address this, We propose the Frequency Recalibration Method (FreRec) involving a two-phase manipulation: first, Statistic High-frequency Replacement (SHR) replaces the high-frequency components of synthetic images with the average from real images, achieving initial alignment; second, Reconstructive High-frequency Mapping (RHM) further mapping the high-frequency distribution to that of real images and restoring image quality through learnable unidirectional reconstruction. Together, these steps provide a coarse-to-fine calibration of frequency distributions for AI-synthesized medical images.

\begin{figure}
    \centering
    \includegraphics[width=1\linewidth]{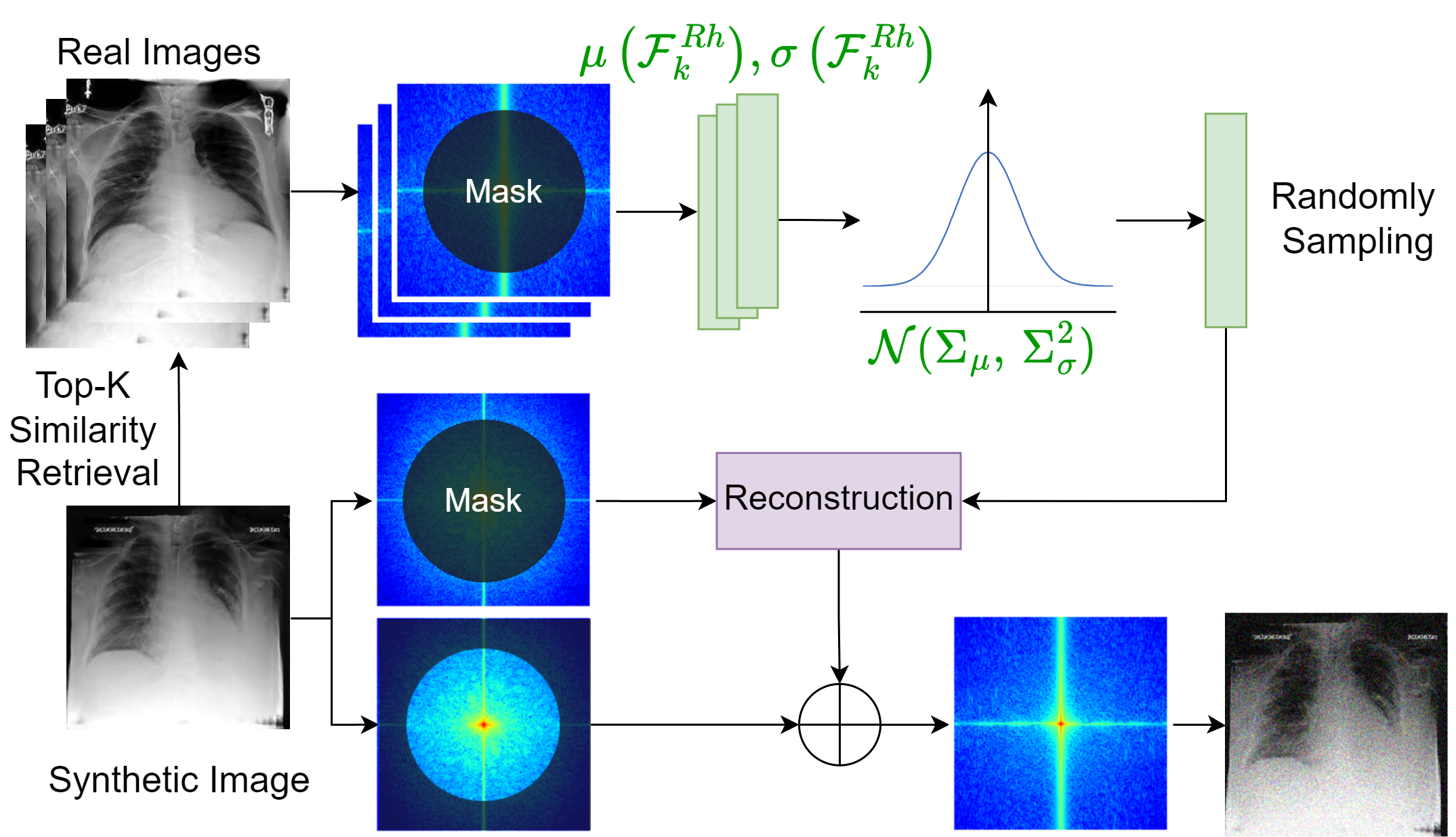}
    \caption{The workflow of Statistic High-frequency Replacement.}
    \label{fig:SHR}
\end{figure}

\subsection{Statistic High-frequency Replacement}
Since AI-synthesized samples implicitly differ from real samples in higher-frequency components \cite{durall2020watch}, a rough distributional alignment can be achieved via replacing the high-frequency components of AI samples with the counterparts of real samples by statistic. 

Let $x_i^{S} \in \mathbb{R}^{N \times N \times C}$ be a synthesized image, its Fourier transformation can be denoted as $\mathcal{F}\left(x_i^{S}\right)(u, v, c)=\sum_{h=0}^{N-1} \sum_{w=0}^{N-1} x_i(h, w, c) e^{-j 2 \pi\left(\frac{h}{N} u+\frac{w}{N} v\right)}$. A binary mask $\mathcal{M} \in \mathcal{R}^{r \times r}$, whose values are zero except for the center region with a fixed ratio $r$, is applied to the centered Fourier spectrum to separated it into the low-frequency component $\mathcal{F}^l\left(x_i^{S}\right)$ and the high-frequency component $\mathcal{F}^h\left(x_i^{S}\right)$. The same operation can be applied to a real medical image $x_i^{R}$ for $\mathcal{F}^l\left(x_i^{R}\right)$ and $\mathcal{F}^h\left(x_i^{R}\right)$, and on this basis the high-frequency components can be exchanged.   

However, one-to-one replacement is prone to randomness that fails to ensure distributional alignment. Therefore, as shown in Fig. \ref{fig:SHR}, for each synthetic image, we select a batch of real images for a statistic-based replacement. Additionally, to minimize perturbation from rough alignment and preserve the semantic features of the modified synthetic image, we retrieve the top-{K} real images $\{x_k^{R}\}_{k=1}^{K}$ that are similar as $x_i^{S}$ by Structural Similarity (SSIM) score for replacement. Then, given $\{x_k^{R}\}$, the channel-wise mean and standard deviation of high-frequency spectral distribution are computed as (for simplicity, we denote $\mathcal{F}^h\left(x_k^{R}\right)$ as $\mathcal{F}^{Rh}_{k}$):

\begin{equation}
\begin{gathered}
\mu\left(\mathcal{F}^{Rh}_{k}\right)=\frac{1}{H^2} \sum_{u=1}^H \sum_{v=1}^H \mathcal{F}^{Rh}_{k}(u, v, c) \\
\sigma\left(\mathcal{F}^{Rh}_{k}\right)=\frac{1}{H^2} \sum_{u=1}^H \sum_{v=1}^H\sqrt{\left[\mathcal{F}^{Rh}_{k}(u, v, c)-\mu\left(\mathcal{F}^{Rh}_{k}\right)\right]^2} .
\end{gathered}
\end{equation}

Assuming the frequency components at each spectral band in a batch of independent images follow a Gaussian distribution (see Supplement for justification), the statistical variances are calculated as follows:

\begin{equation}
\begin{aligned}
\Sigma_\mu\left(\mathcal{F}^{Rh}_{k}\right) & =\sqrt{\frac{1}{K}\sum\left[\mu\left(\mathcal{F}^{Rh}_{k}\right)-\mathbb{E}\left[\mu\left(\mathcal{F}^{Rh}_{k}\right)\right]\right]^2} \\
\Sigma_\sigma\left(\mathcal{F}^{Rh}_{k}\right) & =\sqrt{\frac{1}{K}\sum\left[\sigma\left(\mathcal{F}^{Rh}_{k}\right)-\mathbb{E}\left[\sigma\left(\mathcal{F}^{Rh}_{k}\right)\right]\right]^2}.
\end{aligned}
\end{equation}
Then the Gaussian distribution for probabilistic statistics of high-frequency components of $\{x_k^{R}\}_{k=1}^{K}$ can be modeled. With this Gaussian distribution, we can randomly sample new mean $\hat{\mu}$ and standard deviation $\hat{\sigma}$ to modify the original high-frequency component $\mathcal{F}^{Sh}_{i}$ of the synthetic image $x_i^{S}$:

\begin{equation}
\label{eq3}
\begin{aligned}
& \hat{\mu}\left(\mathcal{F}^{Rh}_{k}\right)=\mu\left(\mathcal{F}^{Rh}_{k}\right)+\epsilon_\mu \Sigma_\mu\left(\mathcal{F}^{Rh}_{k}\right), \epsilon_\mu \sim \mathcal{N}(0, 1), \\
& \hat{\sigma}\left(\mathcal{F}^{Rh}_{k}\right)=\sigma\left(\mathcal{F}^{Rh}_{k}\right)+\epsilon_\sigma \Sigma_\mu\left(\mathcal{F}^{Rh}_{k}\right), \epsilon_\sigma \sim \mathcal{N}(0, 1), \\
&
\hat{\mathcal{F}}^{Sh}_{i}=\hat{\sigma}\left(\mathcal{F}^{Rh}_{k}\right)\left(\frac{\mathcal{F}^{Sh}_{i}-\mu\left(\mathcal{F}^{Rh}_{k}\right)}{\sigma\left(\mathcal{F}^{Rh}_{k}\right)}\right)+\hat{\mu}\left(\mathcal{F}^{Rh}_{k}\right) .
\end{aligned}
\end{equation}

By combining the above $\hat{\mathcal{F}}^{Sh}_{i}$ with the original low-frequency component $\mathcal{F}^{Sl}$, followed by the inverse Fourier transformation, a lossy calibrated sample $\hat{x}_i^{S}$ is obtained. 

\subsection{Reconstructive High-frequency Mapping}
\begin{figure}
    \centering
    \includegraphics[width=0.9\linewidth]{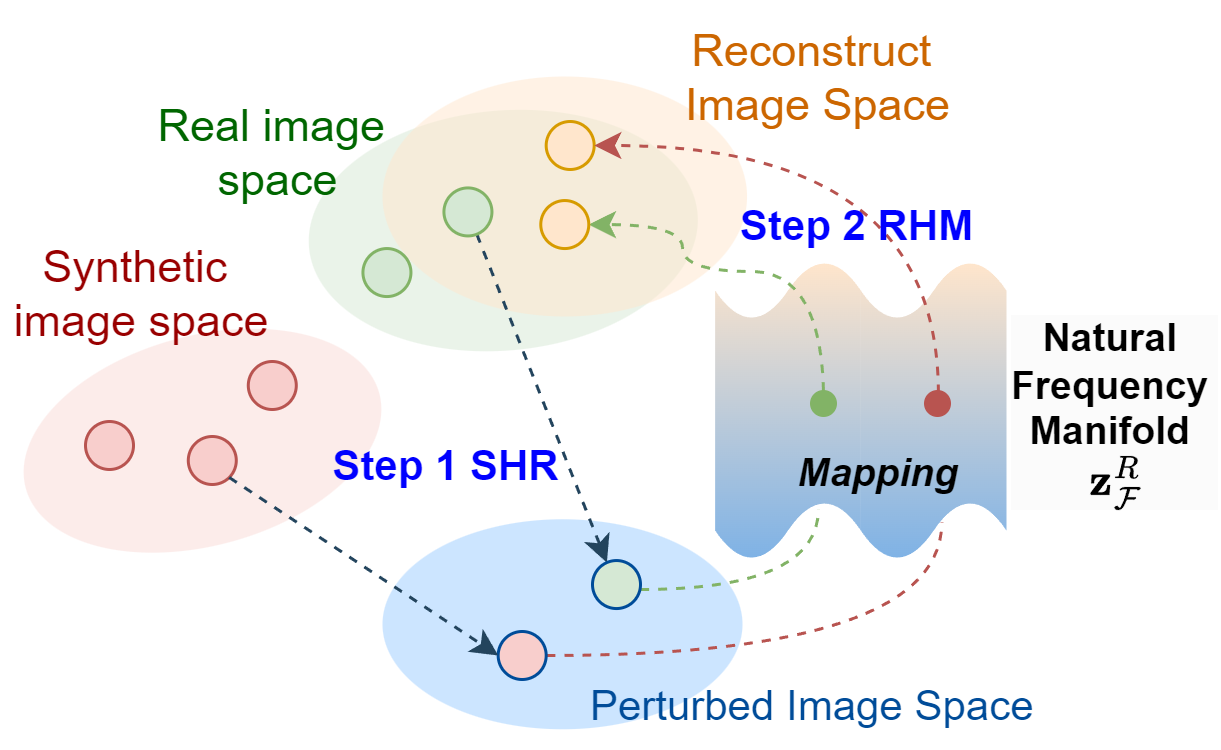}
    \caption{A conceptual explanation of the Reconstructive High-frequency Mapping. Images are first transformed into the same starting space during the initial alignment by SHR. Then the synthetic images can be further calibrated by mapping onto the natural frequency manifold following the same reconstruction path learned from real images.}
    \label{fig:manifold}
\end{figure}

Statistic High-frequency Replacement provides coarse frequency alignment but causes spectral distortion and reduces image quality. Therefore, a further step is needed to restore quality and recover high-frequency details resembling real images from previous perturbed versions. A simple denoising reconstruction of the perturbed synthetic samples via a direct mapping $\hat{x}_i^{S}\xrightarrow{}{x}_i^{S}$ is ineffective, as it reverts back to the original, frequency-misaligned synthetic image. Instead, we address this using unidirectional manifold mapping, i.e., learning a latent natural frequency manifold ($\mathbf{z}_{\mathcal{F}}^{R}$) from real images, and then projecting synthetic images onto it (see Fig. \ref{fig:manifold}). Reconstruction from this shared latent space $\mathbf{z}$ restores high-frequency details, aligning synthetic images with the natural frequency distribution of real images.

\begin{figure*}
    \centering
    \includegraphics[width=0.9\linewidth]{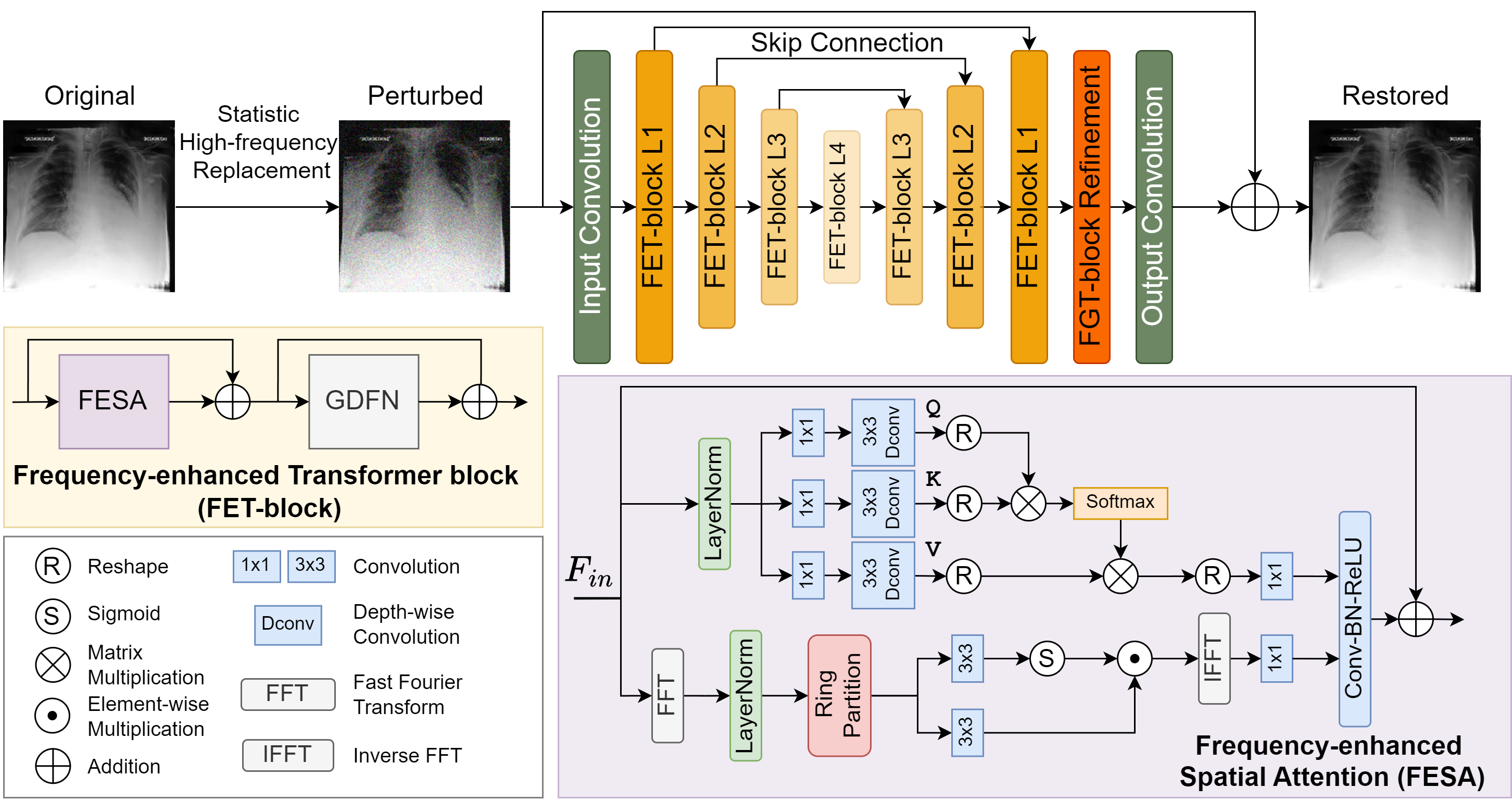}
    \caption{The details of the denoising auto-encoder used in Reconstructive High-frequency Mapping.}
    \label{fig:RHM}
\end{figure*}

To capture the natural frequency manifold $\mathbf{z}_{\mathcal{F}}^{R}$, we first train a denoising auto-encoder $\mathcal{A}: \hat{x}_i^{R}\xrightarrow{}{x}_i^{R}$, where $\hat{x}_i^{R}$ is a perturbed version of ${x}_i^{R}$. Importantly, $\mathcal{A}$ must be \textbf{trained exclusively on real images} to form the accurate frequency reconstruction direction towards the true frequency distribution of real images. Additionally, the input real image $\hat{x}_i^{R}$ is perturbed using the same SHR method (Eq.\ref{eq3}) as well, to ensure a consistent projection path where both real and synthetic images are reconstructed from the same input space, as explained in Fig \ref{fig:manifold}. After training, we apply the well-trained auto-encoder $\mathcal{A}^*$ to synthetic images for high-frequency reconstruction, during which their frequency distribution is calibrated to real images'.

\begin{table*}[htbp]
\setlength{\tabcolsep}{2pt}
\renewcommand{\arraystretch}{1.2}
  \centering
  \caption{Details of datasets and settings.}
  \resizebox{1\textwidth}{!}{
        \setlength{\tabcolsep}{2mm}{
    \begin{tabular}{c|c|c|c|c|c|c|c|c|c}
    \hline
    \multicolumn{2}{c|}{\textbf{Datasets}} & \multicolumn{2}{c|}{\textbf{GenAIs}} & \multicolumn{4}{c|}{\textbf{Training Set}} & \multicolumn{2}{c}{\textbf{Test Set}} \\
    \hline
    \multirow[c]{2}{*}{Task} & \multirow[c]{2}{*}{Dataset} & \multirow[c]{2}{*}{Type} & \multirow[c]{2}{*}{Source} & \multicolumn{2}{c|}{Real } & \multicolumn{2}{c|}{Synthetic } & \multicolumn{2}{c}{Real } \\
\cline{5-10}          &       &       &       & Positive & Negative & Positive & Negative & Positive & Negative \\
    \hline
    Brain tumor MRI & Kaggle-Brain tumor & FastGAN & Pre-trained & 1000  & 1000  & 1500  & 1500  & 300   & 300 \\
    \hline
    DR Fundus image & Kaggle-DR & VC-Diffusion & Pre-trained & 7000  & 20000 & 3000  &  -  & 3000  & 6000 \\
    \hline
    Cardiomegaly X-ray & MIMIC-CXR & StyleGAN3 & From scratch & 800   & 350   & 800   & 1250  & 100   & 100 
    \\
    \hline
    \end{tabular}}}%
  \label{tab:settings}%
\end{table*}%
% \footnotetext[1]{https://www.kaggle.com/datasets/sartajbhuvaji/brain-tumor-classification-mri}
% \footnotetext[2]{https://physionet.org/content/mimic-cxr/2.0.0/}
% \footnotetext[3]{https://www.kaggle.com/competitions/diabetic-retinopathy-detection}

\paragraph{The backbone of $\mathcal{A}$} Designing $\mathcal{A}$ is challenging, as it must accurately capture the frequency and visual details of real images without introducing additional frequency distortion by itself. We address this through both model architecture and loss function design. Specifically, we propose a transformer network based on the Restoration Transformer (Restormer) \cite{zamir2022restormer}, as shown in Fig \ref{fig:RHM}, replacing its transformer blocks with our Frequency-enhanced Transformer blocks (FET-blocks) to better learn high-frequency details. Each FET-block incorporates a novel Frequency-enhanced Spatial Attention (FESA) module and the original Gated-Dconv Feed-forward Network (GDFN) from Restormer. The network contains 2, 4, 6, and 8 FET-blocks in levels 1 through 4, respectively, with two additional FET-blocks in the refinement stage.

FESA is built on a cross-attention mechanism that fuses the global spatial self-attention branch with a local frequency self-attention branch. The global spatial self-attention branch takes the RGB feature as input. The layer-normalized feature $F_{rgb}\in\mathbb{R}^{H\times W\times C}$ is split into query ($Q$), key ($K$) and value ($V$) projections using $1\times 1$ convolutions to aggregate pixel-wise local context followed by $3\times 3$ depth-wise convolutions to encode channel-wise spatial context \cite{zamir2022restormer}. The query and key projections are then reshaped for dot-product, yielding a transposed-attention map of size $\mathbb{R}^{C\times C}$. Overall, the global spatial self-attention can be denoted as: 

\begin{equation}
\mathbf{\hat{F}}_{rgb}=\operatorname{C_{1}}\left(\operatorname{softmax}(Q\cdot K/a)\cdot V\right)
\end{equation}
where $\mathbf{\hat{F}}_{rgb}$ is the output feature map; $Q\in\mathbb{R}^{HW\times C}$, $K\in\mathbb{R}^{C\times HW}$ and $V\in\mathbb{R}^{HW\times C}$ reshaped tensors. $a$ is a learnable scaling parameter to control the magnitude of the dot product. $\operatorname{C_{n}}$ indicates $n\times n$ convolution.

The local frequency self-attention branch enriches the global learning by embedding additional local frequency information into the spatial attention. The RGB feature is transformed into the frequency domain by Fast Fourier Transform (FFT) to obtain its amplitude spectrum as input. To further disentangle frequency components, the average-pooled spectrum feature $\mathbf{F}_{spe}\in\mathbb{R}^{2R\times 2R\times 1}$ is then partitioned into many non-overlapping circular rings with a width of $d$ along the radius dimension, where $d$ defines the range of frequency components. We group the features located in the same circular ring into one channel and obtain $\mathbf{F}_{fre}\in\mathbb{R}^{P\times R/d}$, where $P$ denotes the number of frequency components on each channel. Then, we use two $3\times 3$ convolutions. The first convolution outputs features into a sigmoid function to obtain a local attention map, as sigmoid gates are often used to control the local flow of information without adding a large computational overhead \cite{hochreiter1997long}. The second convolution is used to refine the spectrum features, followed by a point-wise multiplication with the local attention map and inverse FFT. The local frequency self-attention can be denoted as: 
\begin{equation}
\mathbf{\hat{F}}_{fre}=\operatorname{C_{1}}(\operatorname{iFFT}\left(\operatorname{C_{3}}\left(\operatorname{\delta}(\mathbf{F}_{fre})\right)\cdot \operatorname{C_{3}}(\mathbf{F}_{fre})))\right)
\end{equation}

Finally, the global RGB feature and local frequency feature are fused by concatenation and a $1\times 1$-Conv-BN-ReLU block. After that, the original RGB features $\mathbf{F}_{rgb}$ are added, resulting in the refined features: 
\begin{equation}
\mathbf{\hat{F}}_{final}=\mathbf{F}_{rgb}+\operatorname{ReLU-BN-Conv}(\mathbf{\hat{F}}_{rgb}||\mathbf{\hat{F}}_{fre}),
\end{equation}
where $||$ indicated feature concatenation.

% perform a multi-metric evaluation using a novel frequency-level similarity metric FSIM, alongside the image quality metrics PSNR and SSIM. FSIM is defined as:

% \begin{equation}
%     \operatorname{FSIM} := 1 - \sqrt{||\mathcal{F}({x}_i^{R}) - \mathcal{F}\left(\mathcal{A^*}(\hat{x}_i^{R})\right)||^2},
% \end{equation}
% which measures the element-wise Euclidean distance between the Fourier spectra of the original and restored images. We compare several denoising architectures, including CNNs and Transformers, with the state-of-the-art Local Frequency Transformer (LoFormer) \ref{} demonstrating the best performance and selected as the backbone for $\mathcal{A}$, as shown in Fig. \ref{}. This superiority is due to LoFormer’s frequency-specific design, which stacks multiple Local Frequency Transformer Blocks, each employs local frequency-domain channel attention to capture cross-covariance across frequency components and an MLP Gating mechanism to emphasize frequency features while enhancing global learning.

\paragraph{Loss function} In addition to the standard pixel-level similarity loss for image reconstruction, we introduce a spectral similarity loss to further ensure that $\mathcal{A}$ can accurately learn frequency and visual details of real images, resulting in a joint loss function:

\begin{equation}
    \min_{\mathcal{A}} \mathcal{L} := \underbrace{||{x}_i^{R} - \mathcal{A}(\hat{x}_i^{R})||^2}_{\text{Pixel similarity}} + \underbrace{||\mathcal{F}({x}_i^{R}) - \mathcal{F}\left(\mathcal{A}(\hat{x}_i^{R})\right)}_{\text{Frequency similarity}}||^2
\end{equation}

\subsection{Incorporating FreRec with GDA}
Due to the inevitable sub-optimality of $\mathcal{A}^*$, subtle frequency differences may remain between its reconstructions and the original images, causing the reconstructed image space to shift slightly from the real image space (see Fig.\ref{fig:manifold}). Thus, to achieve the best practice, we recommend implementing FreRec as a unified, plug-and-play pre-processing module in both training and inference phases of downstream disease classifiers. That is, all training and testing samples, synthetic and real, should be processed with FreRec to make the final frequency distributions unified throughout the disease classification workflow. 

\section{Experiments}
\begin{figure}
    \centering
    \includegraphics[width=\linewidth]{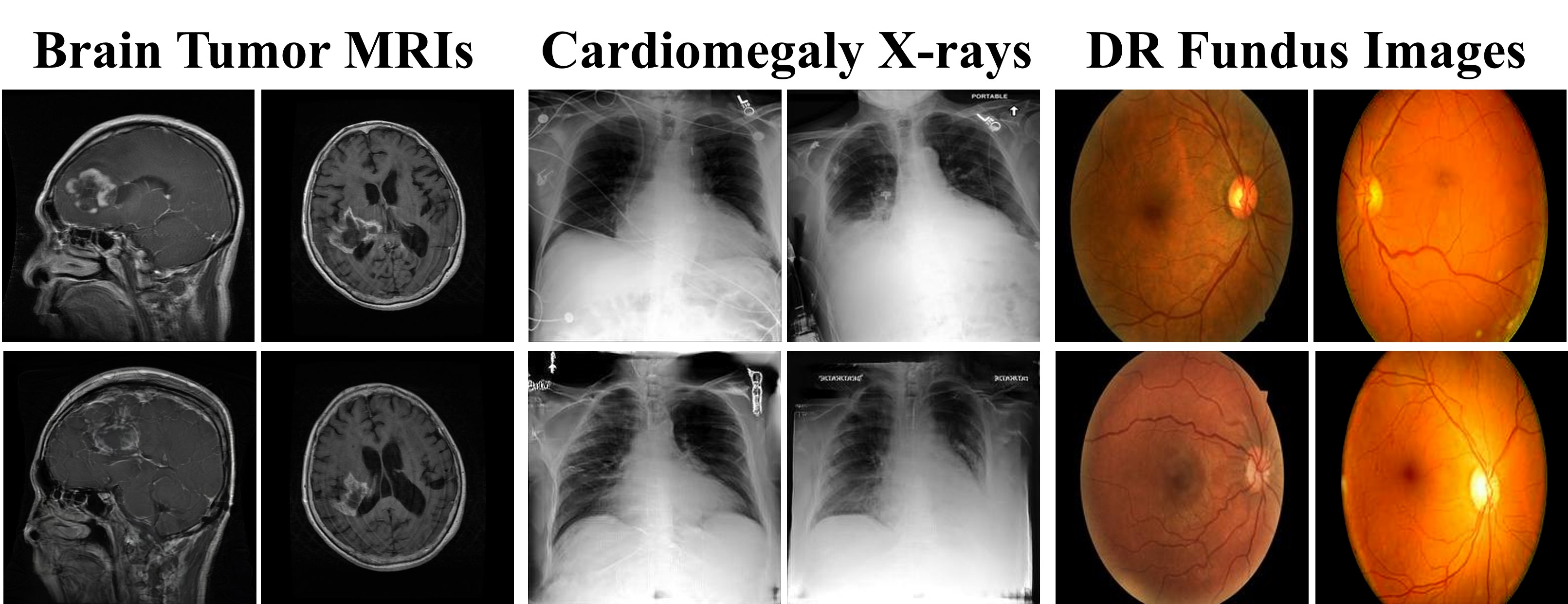}
    \caption{Examples of original real images (above) and synthetic images (bottom) from three datasets. The synthetic images maintain high visual quality and fidelity.}
    \label{fig:syntheticsamples}
\end{figure}

\paragraph{Datasets}
To maintain experimental diversity, we evaluated three medical classification tasks based on different public image datasets: brain tumor detection (brain MRIs) \cite{sartaj_bhuvaji_ankita_kadam_prajakta_bhumkar_sameer_dedge_swati_kanchan_2020}, cardiomegaly diagnosis (chest X-rays) \cite{johnsonMIMICCXRDeidentifiedPublicly2019}, and diabetic retinopathy (DR) classification (fundus photography) \cite{diabetic-retinopathy-detection}. The generative models employed for generative data augmentation (GDA) are varied by model type and source: a pretrained GAN (FastGAN \cite{liu2020towards}) for the synthesis of brain tumor MRI images, a pretrained diffusion model (VC-Diffusion \cite{ilanchezian2023generatingrealisticcounterfactualsretinal}) for DR fundus images, and a from-scratch GAN (StyleGAN3 \cite{Karras2021}) for cardiomegaly X-rays. Both pretrained models have demonstrated GDA effectiveness in their original studies. Figure \ref{fig:syntheticsamples} presents examples of real and synthetic samples from three datasets, while Table \ref{tab:settings} provides a summary of the dataset configurations. We perform GDA to enhance the training sample size while striving to balance the distribution of disease classes as effectively as possible.

\paragraph{Baselines and Settings} We compared no augmentation (RAW), GDA without frequency alignment (GDA), GDA with FreRec (GDA+FreRec). In addition, we also assessed three image processing-based augmentation methods, AutoAug \cite{cubuk2019autoaugment}, Mix-up \cite{Psaroudakis_2022_CVPR}, and Fourier-basis Augmentation (AFA) \cite{10655510}, plus Domain Gap Embeddings (DoGE), a recent domain bias reduction method for AI-synthesized images \cite{10657264}. Different classifier backbones, including ResNet50, DenseNet, and ViT-B-16, were evaluated for disease classification. Classification accuracy (Acc.), F1 and AUC scores were the evaluation metrics, with each classifier evaluated five times and the average results are reported to reduce randomness. The mask ratio $r$ and the sample number $k$ of SHR was set to $0.5$ and $200$ respectively, considering the trade-off between alignment effect, image quality and running time (see Supplement for the decision process).

\paragraph{Disease classification results}
Table \ref{tab:cls} shows the classification results of three diseases under various augmentation strategies. Normal augmentations (AutoAug, Mix-up, and AFA) slightly improve performance over raw training, where AutoAug and AFA are more reliable than Mix-up, which occasionally degrades performance. The plain generative data augmentation, GDA, also suffers from unreliability, sometimes having a negative impact on classifiers despite significantly expanding the training set with synthetic images. For instance, GDA decreases accuracy and F1 scores of cardiomegaly-ResNet50 and all brain tumor classifiers compared to raw training. This observation aligns with prior findings that AI-synthesized samples do not always improve downstream tasks \cite{singh2024synthetic}. In comparison, recalibrating frequency with FreRec (GDA+FreRec) significantly improves the augmentation effect of GDA, making GDA more reliable. GDA+FreRec consistently improves raw classifiers across all tests and metrics, demonstrating its effectiveness as an augmentation strategy. This supports the hypothesis that the frequency distributional gap between real and synthetic samples is a key cause of GDA instability, and closing this gap via frequency calibration improves reliability. In general, GDA+FreRec outperforms baseline augmentation  strategies in all tasks except for DoGE in DR classification. This is because DoGE directly reduces domain bias in synthetic images via source-to-target domain adaptation. However, it heavily relies on a fixed, known synthetic image source, rendering DoGE only applicable to specific generation models. In contrast, FreRec is trained solely on real images, enabling a unidirectional mapping of synthetic images from unknown and arbitrary generation models to real ones, suggesting greater adaptability to unseen synthetic images.

\paragraph{Ablation study} We compare the complete FreRec with sole SHR and sole RHM, with and without FESA (i.e., the original Restormer), in terms of classification performance and reconstruction quality. As shown in Table \ref{tab:ablation}, FreRec consistently outperforms its partial variants in all classification tasks. While RHM without FESA yields the best image quality due to its 1:1 reconstruction, the quality difference between it and FreRec is minor, indicating FreRec achieves frequency alignment at a negligible cost of image quality. Notably, sole SHR does not improve classification performance despite roughly aligning frequency distributions, as the alignment process significantly distorts image quality. Therefore, the RHM step is essential for fine-grained calibration and restoration of image details. Reconstructed synthetic samples can be visualized in the Supplement.

\begin{table*}[htbp]
\setlength{\tabcolsep}{6pt}
\renewcommand{\arraystretch}{1.15}
  \centering
  \caption{Disease classification results in three datasets using different augmentation strategies. \textbf{Bold} and \underline{Underline} indicates the best and second best results in each group, respectively. Card = cardiomegaly; DR = Diabetic Retinopathy; BT = Brain Tumor.}
  \resizebox{1\textwidth}{!}{
        \setlength{\tabcolsep}{5.2mm}{
    \begin{tabular}{c|c|c|c|c|c|c|c|c|c|c}
    \hline
    \multicolumn{1}{c}{\multirow{2}[4]{*}{}} &       & \multicolumn{3}{c|}{DenseNet} & \multicolumn{3}{c|}{ResNet50} & \multicolumn{3}{c}{VIT} \\
\cline{3-11}    \multicolumn{1}{c}{} &       & AUC   & Acc. & F1 & AUC   & Acc. & F1 & AUC   & Acc. & F1  \\
    \hline
    \multirow{8}[16]{*}{Card} & RAW   & 0.842 & 0.803 & 0.787 & 0.834 & 0.792 & 0.788 & 0.832 & 0.794 & 0.779 \\
         & AutoAug & 0.853 & 0.814 & 0.798 & 0.847 & 0.803 & 0.797 & 0.843 & 0.804 & 0.787 \\
         & Mix-up & 0.833 & 0.812 & 0.795 & 0.828 & 0.804 & 0.787 & 0.823 & 0.808 & 0.792 \\
         & AFA   & 0.855 & 0.808 & 0.800 & 0.831 & 0.822 & 0.815 & 0.859 & 0.819 & 0.817 \\
         & DoGE  & 0.880 & 0.833 & 0.833 & 0.869 & 0.821 & \textbf{0.836} & 0.851 & 0.818 & 0.822 \\
         & GDA   & 0.871 & 0.803 & 0.804 & 0.834 & 0.782 & 0.783 & 0.848 & 0.813 & 0.814 \\
         & GDA+FreRec & \textbf{0.899} & \textbf{0.848} & \textbf{0.834} & \textbf{0.888} & \textbf{0.832} & \underline{0.834} & \textbf{0.888} & \textbf{0.838} & \textbf{0.832} \\
    \hline
    \multirow{8}[16]{*}{DR} & RAW   & 0.840 & 0.803 & 0.793 & 0.843 & 0.792 & 0.783 & 0.834 & 0.794 & 0.782 \\
         & AutoAug & 0.857 & 0.821 & 0.809 & 0.861 & 0.808 & 0.799 & 0.848 & 0.811 & 0.798 \\
         & Mix-up & 0.852 & 0.791 & 0.804 & 0.831 & 0.780  & 0.794 & 0.847 & 0.783 & 0.793 \\
         & AFA   & 0.861 & 0.805 & 0.810 & 0.855 & 0.801 & 0.788 & 0.837 & 0.801 & 0.800 \\
         & DoGE  & \textbf{0.880} & \textbf{0.841} & 0.816 & 0.871 & \textbf{0.825} & \textbf{0.819} & \textbf{0.862} & 0.811 & \textbf{0.810} \\
         & GDA   & 0.863 & 0.813 & 0.804 & 0.848 & 0.803 & 0.792 & 0.834 & 0.783 & 0.773 \\
         & GDA+FreRec & \underline{0.879} & \underline{0.834} & \textbf{0.823} & \textbf{0.878} & \underline{0.823} & \underline{0.813} & \underline{0.852} & \textbf{0.813} & \underline{0.803} \\
    \hline
    \multirow{8}[16]{*}{BT} & RAW   & 0.840 & 0.742 & 0.733 & 0.793 & 0.783 & 0.772 & 0.753 & 0.723 & 0.713 \\
         & AutoAug & 0.833  & 0.759  & 0.740  & 0.811  & 0.799  & 0.779  & 0.771  & 0.731  & 0.729  \\
         & Mix-up & 0.853  & 0.731  & 0.744  & 0.782  & 0.794  & 0.761  & 0.764  & 0.734  & 0.702  \\
         & AFA   & 0.841  & 0.824  & 0.730 & 0.807  & 0.814  & 0.766  & 0.771  & 0.728  & 0.739  \\
         & DoGE  & \textbf{0.859} & 0.833  & 0.742  & 0.811  & 0.822  & 0.780  & 0.785  & 0.733  & 0.747  \\
         & GDA   & 0.794 & 0.732 & 0.713 & 0.783 & 0.723 & 0.723 & 0.758 & 0.732 & 0.743 \\
         & GDA+FreRec & \underline{0.855} & \textbf{0.834} & \textbf{0.753} & \textbf{0.843} & \textbf{0.843} & \textbf{0.783} & \textbf{0.787} & \textbf{0.737} & \textbf{0.752} \\
    \hline
    \end{tabular}}}%
  \label{tab:cls}%
\end{table*}%

% Table generated by Excel2LaTeX from sheet 'Sheet4'
\begin{table}[htbp]
  \centering
  \caption{Classification performances and image quality of different frequency calibration methods.}
  \renewcommand{\arraystretch}{1.1}
  \resizebox{\linewidth}{!}{
        \setlength{\tabcolsep}{1.3mm}{
    \begin{tabular}{l|l|ll|ll|ll|ll}
    \hline
    \multicolumn{1}{l}{} &       & \multicolumn{2}{c|}{AUC} & \multicolumn{2}{c|}{Acc.} & \multicolumn{2}{c|}{PSNR} & \multicolumn{2}{c}{SSIM} \\
    \hline
    \multirow{4}[2]{*}{Card} & SHR   & 0.81  &       & 0.79  &       & 25.10 &       & 0.76  &  \\
          & RHM w.o. FESA & 0.85  & \textcolor{green}{+0.04}  & 0.79  & \textcolor{green}{+0.01}  & 36.44 & \textcolor{green}{+11.34} & 0.98  & \textcolor{green}{+0.21} \\
          & RHM w. FESA & 0.87  & \textcolor{green}{+0.02}  & 0.82  & \textcolor{green}{+0.03}  & 35.51 & \textcolor{red}{-0.93} & 0.96  & \textcolor{red}{-0.02} \\
          & FreRec & 0.89  & \textcolor{green}{+0.03}  & 0.84  & \textcolor{green}{+0.02}  & 35.62 & \textcolor{green}{+0.11}  & 0.95  & \textcolor{green}{+0.00} \\
    \hline
    \multirow{4}[2]{*}{DR} & SHR   & 0.81  &       & 0.77  &       & 25.22 &       & 0.76  &  \\
          & RHM w.o. FESA & 0.85  & \textcolor{green}{+0.04}  & 0.81  & \textcolor{green}{+0.05}  & 36.33 & \textcolor{green}{+11.11} & 0.97  & \textcolor{green}{+0.22} \\
          & RHM w. FESA & 0.86  & \textcolor{green}{+0.01}  & 0.82  & \textcolor{green}{+0.01}  & 34.23 & \textcolor{red}{-2.10} & 0.94  & \textcolor{red}{-0.03} \\
          & FreRec & 0.87  & \textcolor{green}{+0.01}  & 0.82  & \textcolor{green}{+0.01}  & 34.29 & \textcolor{green}{+0.06}  & 0.95  & \textcolor{green}{+0.01} \\
    \hline
    \multirow{4}[2]{*}{BT} & SHR   & 0.78  &       & 0.73  &       & 25.95 &       & 0.77  &  \\
          & RHM w.o. FESA & 0.79  & \textcolor{green}{+0.01}  & 0.75  & \textcolor{green}{+0.02}  & 41.23 & \textcolor{green}{+15.28} & 0.99  & \textcolor{green}{+0.22} \\
          & RHM w. FESA & 0.80  & \textcolor{green}{+0.01}  & 0.80  & \textcolor{green}{+0.05}  & 40.21 & \textcolor{red}{-1.02} & 0.98  & \textcolor{red}{-0.01} \\
          & FreRec & 0.83  & \textcolor{green}{+0.02}  & 0.81  & \textcolor{green}{+0.01}  & 41.15 & \textcolor{green}{+0.94}  & 0.98  & \textcolor{green}{+0.00} \\
    \hline
    \end{tabular}%
  \label{tab:ablation}}}%
\end{table}%

\subsection{Frequency recalibration effectiveness}
\subsubsection{Distribution visualization} 
To confirm the effectiveness of FreRec in aligning frequency distributions of synthetic and real images, we present a visualization of their frequency distributions before and after recalibration. The frequency distribution is computed as a one-dimensional profile by azimuthally integrating the spectral magnitudes over radial frequencies $\theta$ \cite{durall2020watch}. The details of computing the one-dimensional profile can be found in Supplement.

% For an image $I \in \mathbb{R}^{N \times N}$, the one-dimensional frequency profile is:

% \begin{equation}
% \label{eq:fd}
% \operatorname{FD}\left(r_{k}\right)=B_0\int_{0}^{2 \pi}|\mathcal{F}(r_{k}, \theta)| \mathrm{~d} \theta, \quad k=0,...,\frac{N}{2}-1, 
% \end{equation}
% where $B_0$ is a normalization constant, $(r_{k}, \theta)$ is the polar coordinate transformed from $(u,v)$: $r_{k}=\sqrt{u^{2}+v^{2}}$, $\theta=\atantwo\left(v, u\right)$.
% For ease we normalize $r_k$ into the range of $[0, 1]$ using the factor $\frac{1}{\sqrt{\frac{1}{2} N^2}}$, and use a log-scaled spectrum instead of the raw spectrum.

\begin{figure}
    \centering
    \includegraphics[width=\linewidth]{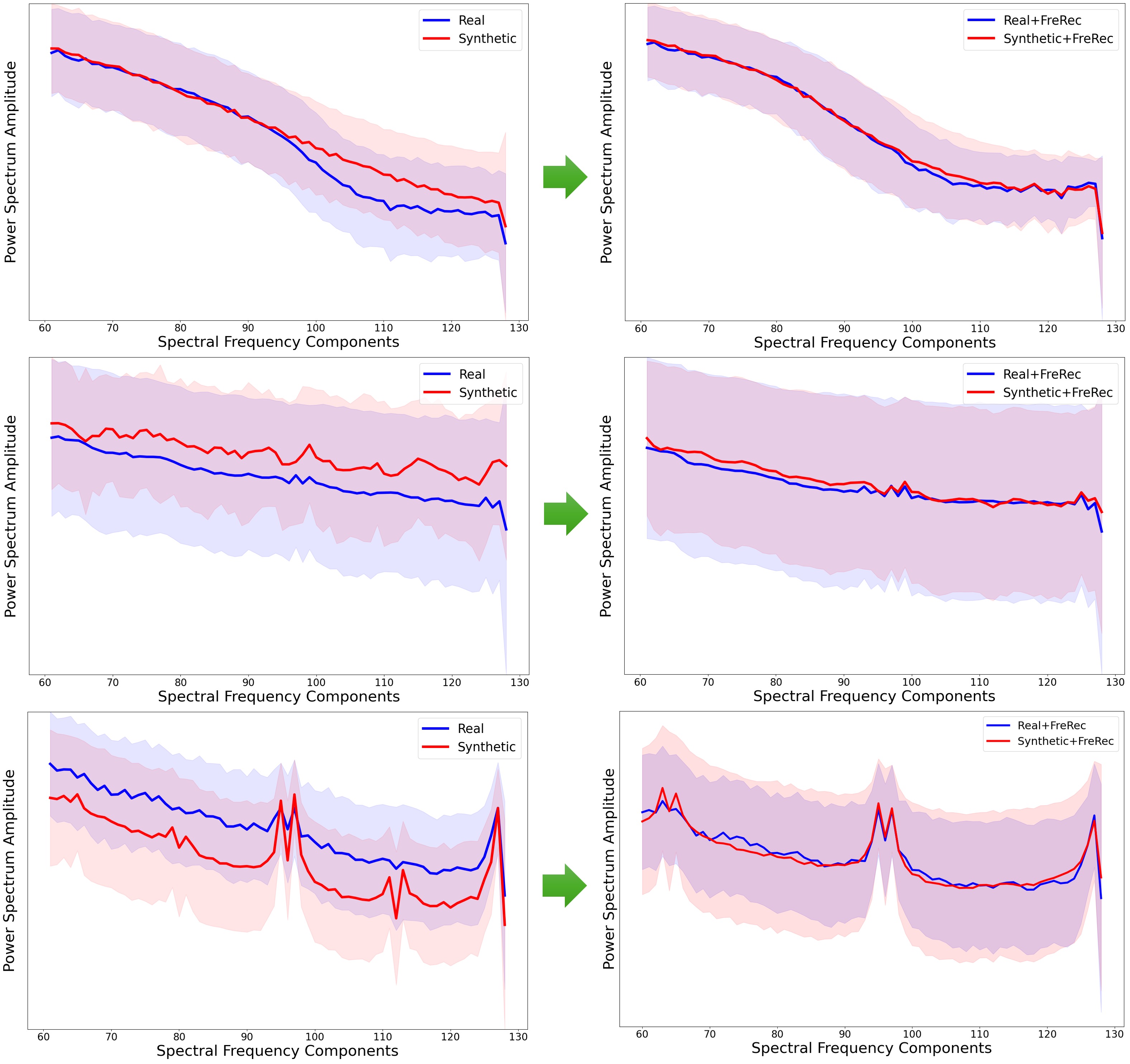}
    \caption{Frequency distributions of real and synthetic images before (the left column) and after (the right column) frequency recalibration in Brain Tumor MRI, Cardiomegaly X-ray, and DR Fundus image datasets (from top to bottom).}
    \label{fig:fredis}
\end{figure}

Figure \ref{fig:fredis} illustrates the average frequency distributional comparisons across three datasets. Prior to applying FreRec, substantial frequency discrepancies were observed between real and synthetic images in all datasets, despite the synthetic images spanning three distinct modalities and being generated by different AI models. After recalibration, these distributional gaps were dynamically reduced across all datasets, leading to improved alignment in frequency distributions. Notably, the alignments in the Brain Tumor MRI and Cardiomegaly X-ray datasets are more thorough compared to those in DR Fundus images. This discrepancy may be attributed to the fact that DR Fundus images are color photographs, whereas the others are grayscale. Color images inherently contain richer pixel information, making the learning of Reconstructive High-frequency Mapping more challenging for the denoising auto-encoder $\mathcal{A}$.

\begin{figure}
    \centering
    \includegraphics[width=\linewidth]{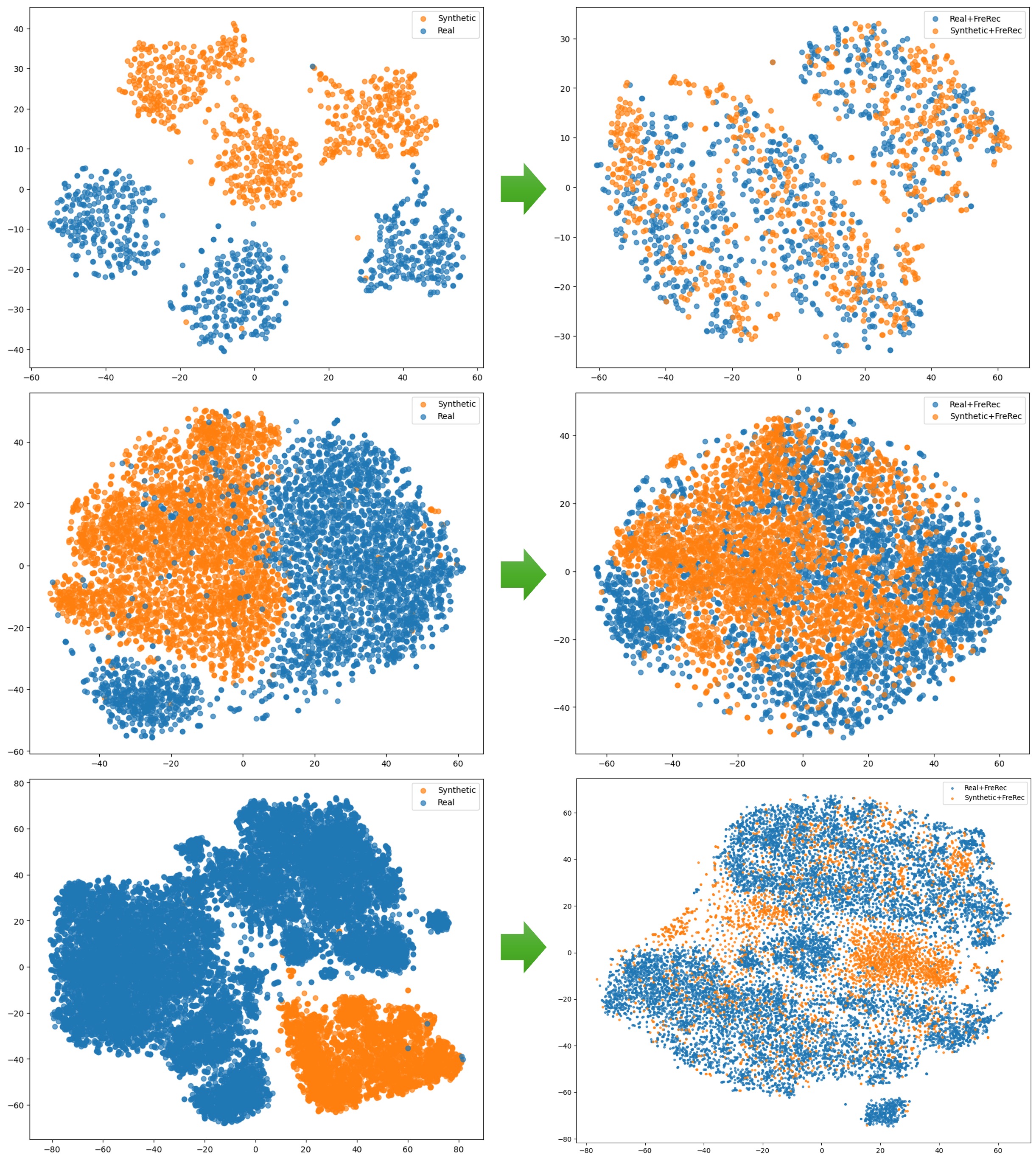}
    \caption{Feature-space visualization of real and synthetic images before (the left column) and after (the right column) frequency recalibration in Brain Tumor MRI, Cardiomegaly X-ray, and DR Fundus image datasets (from top to bottom). Orange dots: Real samples; Blur dots: synthetic samples.}
    \label{fig:tsne}
\end{figure}
\subsubsection{Feature-space visualization} To verify that frequency discrepancies between synthetic and real images contribute as a main cause for GDA's instability in downstream disease classification and that frequency alignment improves its reliability in enriching useful features, we provide a feature-space visualization using T-SNE. For each disease, a ResNet50 classifier is independently trained on $80\%$ of the real images from the training set described in Table \ref{tab:settings}. The classification head is then removed, turning the classifier into a pre-trained feature encoder to extract features from the remaining $20\%$ of real images and an equal number of randomly selected synthetic images. To ensure unbiased evaluation, this process is performed without regard to class labels. The extracted features are visualized in a two-dimensional space using T-SNE.

Figure \ref{fig:tsne} presents the clustering results. In the left column, the features of synthetic and real samples are clearly separated across all datasets, indicating a domain bias in the synthetic samples that can compromise GDA reliability. In the right column, frequency recalibration shifts the synthetic features closer to the real ones. For the Brain Tumor MRI and Cardiomegaly X-ray datasets, the synthetic and real features completely overlap, eliminating domain bias. In the DR dataset, the shift is not complete, likely due to color images have richer pixel information that may complicate the high-frequency reconstruction and induce other contextual biases that frequency recalibration cannot fully address.

% \subsection{$\mathcal{A}$ backbone selection}

\section{Conclusion}
Generative data augmentation is widely used for medical image classification tasks. It can synthesize realistic medical image samples to complement the original dataset. However, the reliability of AI-synthesized samples should be carefully investigated to avoid any domain bias and negative features brought by generative AI models. In this study, we explored this problem from a frequency perspective. We identified that the frequency misalignment between real and synthetic images is a main cause of the instability of GDA. To address this issue, we proposed a novel frequency recalibration method, which consists of two steps, Statistical High-frequency Replacement and Reconstructive High-frequency Mapping, to formulate a coarse-to-fine alignment of the frequency distributions between real and synthetic images. Our extensive experiments performed on various datasets and on different generative AI models demonstrate that the calibrated synthetic samples have a much closer distribution to the real ones, and can significantly improve downstream medical image classifications.

\section{Acknowledgments}
This work was supported by the National Natural Science Foundation of China (Grant No. 62402009), and the Science and Technology Development Fund of Macao under Grants 0002/2024/RIA1 and 0013-2024-ITP1.

% \bigskip
% \noindent Thank you for reading these instructions carefully. We look forward to receiving your electronic files!

\bibliography{aaai2026}

\newpage

\input{supp/supplement-arxiv}
% \input{supp/ReproducibilityChecklist}
\end{document}

%% file: supp/supplement-arxiv.tex
\section{Appendix}
\subsection{Computing 1D Frequency Distribution}
We visualize the frequency distributions before and after recalibration to verify the effectiveness of FreRec in aligning frequency distributions of synthetic and real images. The frequency distribution is computed as a one-dimensional profile by azimuthally integrating the spectral magnitudes over radial frequencies $\theta$ \cite{durall2020watch}. 

For an image $I \in \mathbb{R}^{N \times N}$, its discrete Fourier Transform can be denoted as

\begin{equation}
    \mathcal{F}\left(I\right)(u, v, c)=\sum_{h=0}^{N-1} \sum_{w=0}^{N-1} I(h, w) e^{-j 2 \pi\left(\frac{h}{N} u+\frac{w}{N} v\right)}
\end{equation}. Then its one-dimensional frequency profile is:

\begin{equation}
\label{eq:fd}
\operatorname{FD}\left(r_{k}\right)=B_0\int_{0}^{2 \pi}|\mathcal{F}(r_{k}, \theta)| \mathrm{~d} \theta, \quad k=0,...,\frac{N}{2}-1, 
\end{equation}
where $B_0$ is a normalization constant, $(r_{k}, \theta)$ is the polar coordinate transformed from $(u,v)$: $r_{k}=\sqrt{u^{2}+v^{2}}$, $\theta=\atantwo\left(v, u\right)$.
For ease we normalize $r_k$ into the range of $[0, 1]$ using the factor $\frac{1}{\sqrt{\frac{1}{2} N^2}}$, and use a log-scaled spectrum instead of the raw spectrum. Sup-Fig. \ref{fig:supfig1} illustrates a schematic example of this operation in the brain MRI dataset:

\begin{figure}[!htbp]
    \centering
    \includegraphics[width=\linewidth]{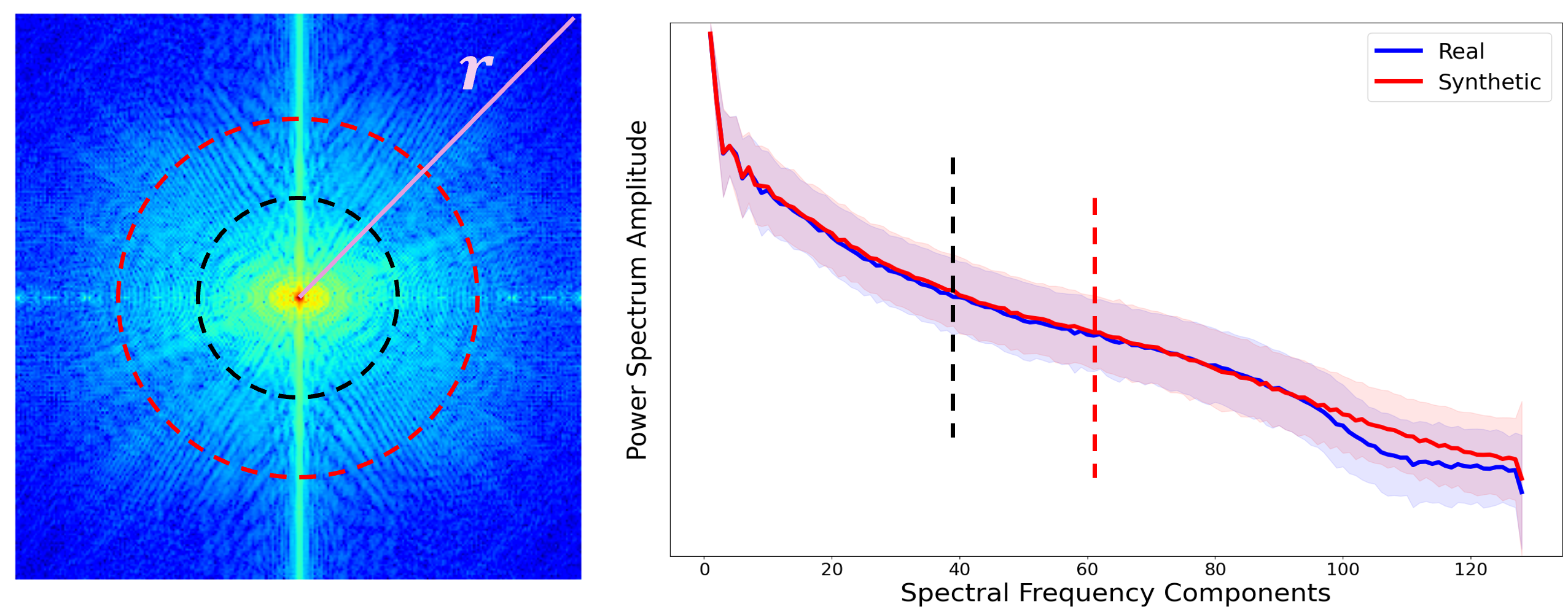}
    \caption{Schematic example of one-dimensional frequency profile computation in the brain MRI dataset}
    \label{fig:supfig1}
\end{figure}

We observed that the frequency distribution of uncalibrated raw synthetic images differs markedly from real images for higher-frequency components ($k > \frac{N}{4}$), but shows minimal discrepancy for lower frequencies ($k \leq \frac{N}{4}$). Therefore, in the main paper, we limit the visualizations of the frequency distribution to $k > \frac{N}{4}$ for a better view. 

\subsection{Hyper-parameter Selection}
\begin{figure*}[!htp]
    \centering
    \includegraphics[width=0.9\textwidth]{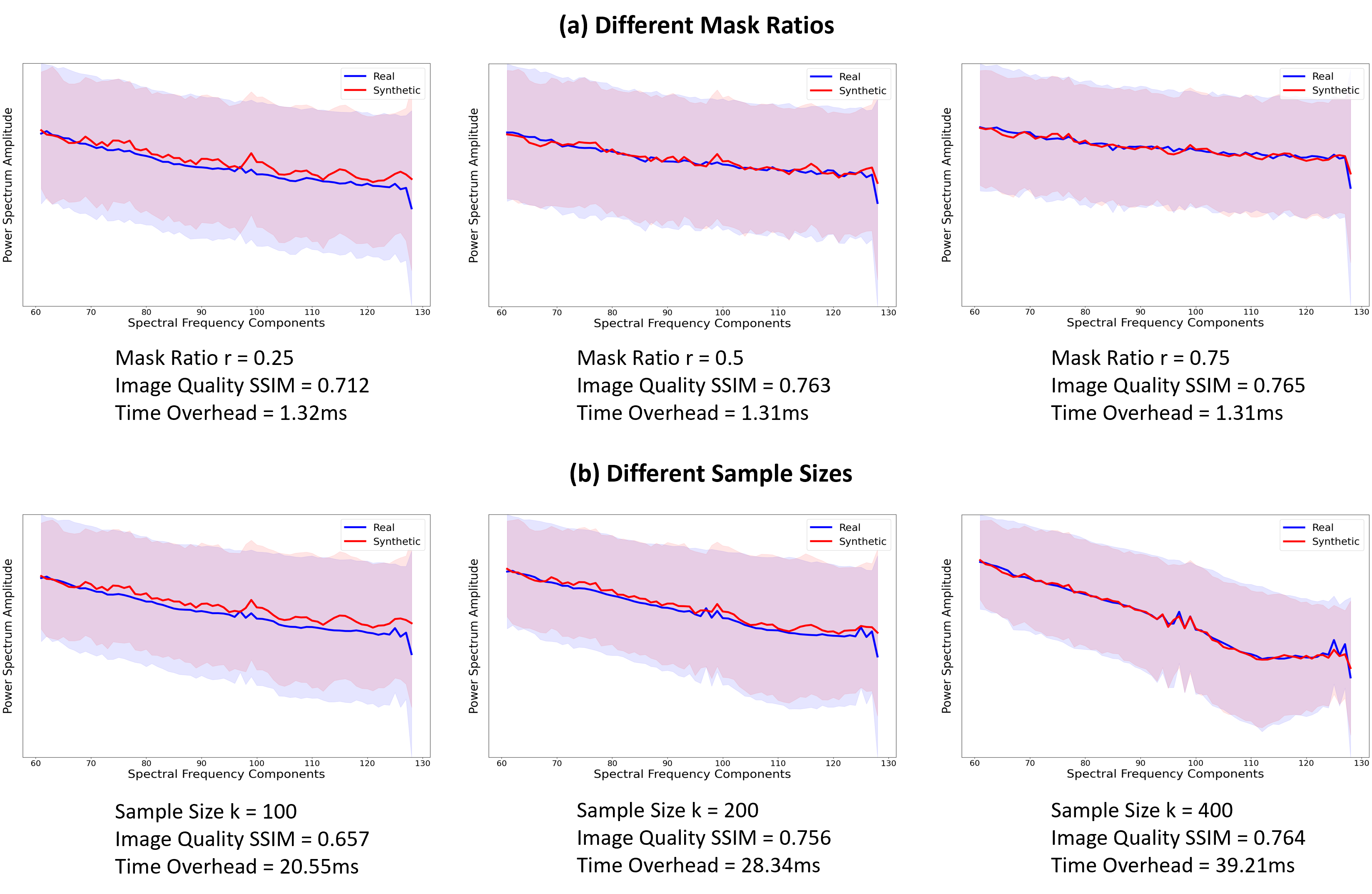}
    \caption{The alignment effect, image quality and running time using different mask ratios and sample sizes.}
    \label{fig:supfig2}
\end{figure*}

The Statistic High-frequency Replacement (SHR) phase of FreRec involves two adjustable parameters, the ratio $r$ of the mask for frequency decomposition, and the sample size $k$ when retrieving the top-{K} real images $\{x_k^{R}\}_{k=1}^{K}$ that are similar as $x_i^{S}$ for replacement. We determine the parameters considering the trade-off between
alignment effect, image quality and running time. Sup-Fig \ref{fig:supfig2} shows the assessment results averaged over three datasets. 

Finally, the mask ratio and sample size are set to 0.5 and 200, respectively.

\subsection{Justification of Assumptions behind FreRec}
\begin{figure*}[!htbp]
    \centering
    \includegraphics[width=0.9\textwidth]{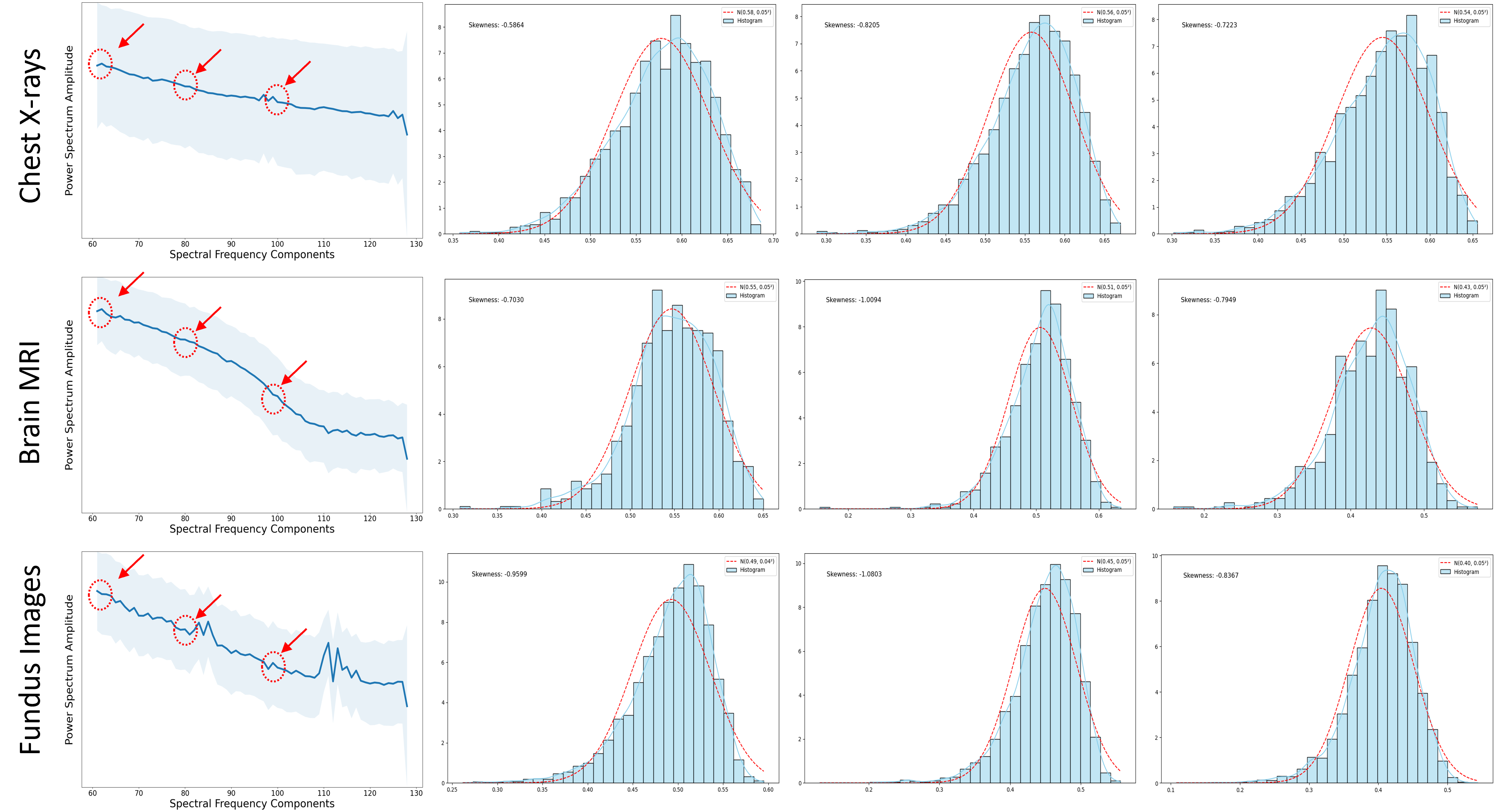}
    \caption{The spectral amplitude histograms at bands $k = 60, 80$, and $100$ approximate a Gaussian distribution.}
    \label{fig:supfig3}
\end{figure*}

The SHR phase of FreRec assumes that frequency components at the same spectral band across independent images follow a Gaussian distribution. To validate this, we randomly sample $1,000$ images from each dataset and plot histograms of spectral amplitudes at bands $k = 60, 80$, and $100$. As shown in Sup-Fig. \ref{fig:supfig3}, all histograms closely resemble a Gaussian distribution, with an average skewness of $-0.835$, supporting the assumption.

\begin{figure}[!htbp]
    \centering
    \includegraphics[width=\linewidth]{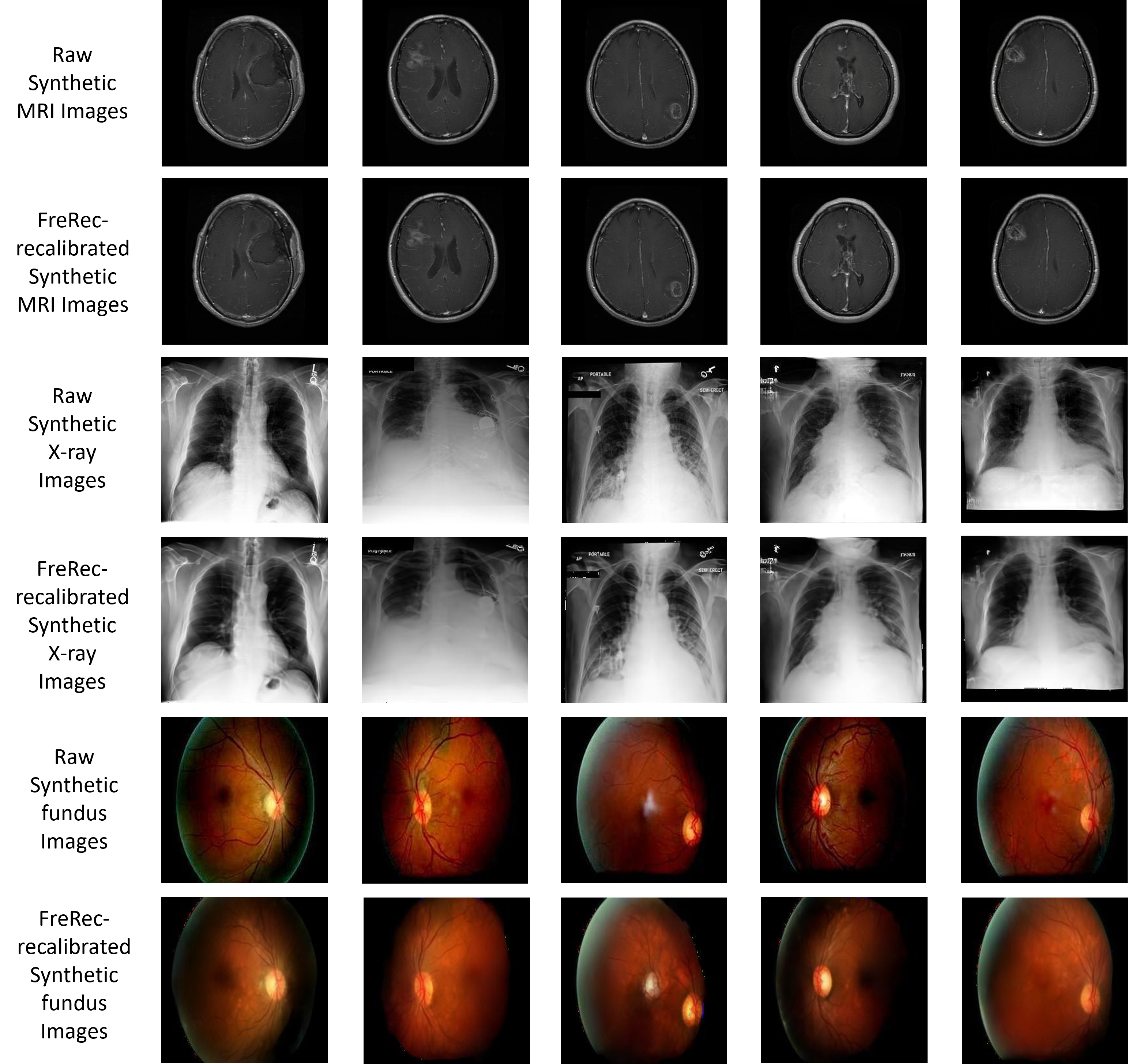}
    \caption{Examples of raw and FreRec-recalibrated synthetic images from three datasets.}
    \label{fig:supfig4}
\end{figure}

\subsection{Time overhead of FreRec}
Since FreRec is a standalone post-processing step compatible with any generative model, and can be implemented as a plug-and-play and cost-effective module that integrates into common medical AI pipelines, it is essential to evaluate its inference time to ensure it adds minimal computational overhead in implementation.

Sup-Table \ref{tab:suptable1} presents the inference time and reconstruction quality of FreRec across three datasets, alongside a comparison with Domain Gap Embeddings (DoGE), a recent domain bias reduction method \cite{10657264}. All tests are performed on a NVIDIA GTX 4090 server. The inference times per image are 16.17 ms for chest X-ray, 15.04 ms for fundus images, and 17.63 ms for brain MRI, all clinically acceptable considering the high reconstruction quality. And the the inference times are less than or comparable to DoGE.

% Table generated by Excel2LaTeX from sheet 'Sheet5'
\begin{table}[!htbp]
\renewcommand{\arraystretch}{1.2}
  \centering
  \caption{The inference time and reconstruction quality of FreRec in three datasets.}
  \resizebox{1\linewidth}{!}{
        \setlength{\tabcolsep}{3mm}{
    \begin{tabular}{l|l|l|l|l}
    \hline
    \multicolumn{1}{l}{} &       & \multicolumn{1}{c|}{Inference Time} & \multicolumn{1}{c|}{SSIM} & \multicolumn{1}{c}{PSNR} \\
    \hline
    \multirow{2}[2]{*}{Card} & DoGE  & 33.21 ms & N/A   & N/A \\
          & FreRec & 16.17 ms & 0.953 & 35.62 \\
    \hline
    \multirow{2}[1]{*}{DR} & DoGE  & 36.14 ms & N/A   & N/A \\
          & FreRec & 15.04 ms & 0.948 & 34.29 \\
    \hline
    \multirow{2}[1]{*}{BT} & DoGE  & 33.78 ms & N/A   & N/A \\
          & FreRec & 17.63 ms & 0.981 & 41.15 \\
    \hline
    \end{tabular}}}%
  \label{tab:suptable1}%
\end{table}%

\subsection{Visualization of Recalibrated Synthetic Images}
Sup-Fig. \ref{fig:supfig4} shows examples of raw and FreRec-recalibrated synthetic images from three datasets. Although FreRec aligns the frequency distributions of synthetic images with real ones, the images retain high visual quality comparable to the originals. All pathological features essential for disease classification are clearly preserved, indicating that FreRec effectively balances frequency calibration and image quality.

\subsection{Generalization to Non-medical Domains}
We also explore whether FreRec can be applied to non-medical images by testing it on the CIFAR-10 dataset. The generative model for GDA is a StyleGAN2 \cite{karras2020analyzing} pretrained by the StudioGAN project \cite{kang2023studiogan}. We evaluate a multi-classification task on the $Automobile$, $Ship$, and $Truck$ categories. Each training class contains 5,000 real and 5,000 synthetic images, while each test class has 1,000 real images. Three classifiers, including ResNet18, ResNet50, and DenseNet, are assessed.

Sup-Table \ref{tab:suptable2} summarizes classification accuracy for models trained on real images only (RAW), with generative data augmentation (GDA), and with FreRec-processed GDA. Unlike the unstable performances in the medical domain, GDA improves classification performance more stably in natural images, likely because semantic features in natural images are more prominent and less sensitive to frequency misalignment than subtle pathological features in medical images. Thus, reducing frequency bias in GDA is especially important for medical data. Nevertheless, FreRec further improves GDA, raising accuracy from about $0.945$ to $0.980$, verifying its effectiveness in non-medical domains. 

% Table generated by Excel2LaTeX from sheet 'Sheet6'
\begin{table}[htbp]
  \centering
  \renewcommand{\arraystretch}{1.2}
  \caption{Accuracy of the $Automobile$, $Ship$, and $Truck$ classification in the CIFAR-10 dataset.}
  \resizebox{1\linewidth}{!}{
        \setlength{\tabcolsep}{3mm}{
    \begin{tabular}{c|c|c|c|c}
    \hline
    \multicolumn{1}{c}{\multirow{2}[4]{*}{}} &       & ResNet18 & ResNet50 & DenseNet \\
\cline{3-5}    \multicolumn{1}{c}{} &       & Accuracy & Accuracy & Accuracy \\
    \hline
    \multirow{3}[6]{*}{CIFAR-10} & RAW   & 0.921 & 0.924 & 0.928 \\
\cline{2-5}          & GDA   & 0.941 & 0.946 & 0.949 \\
\cline{2-5}          & FreRec & 0.973 & 0.978 & 0.981 \\
    \hline
    \end{tabular}}}%
  \label{tab:suptable2}%
\end{table}%

% \bibliography{aaai2026}


% \end{document}

%% file: aaai2026.bib
@article{litjens2017survey,
  title={A survey on deep learning in medical image analysis},
  author={Litjens, Geert and Kooi, Thijs and Bejnordi, Babak Ehteshami and Setio, Arnaud Arindra Adiyoso and Ciompi, Francesco and Ghafoorian, Mohsen and Van Der Laak, Jeroen Awm and Van Ginneken, Bram and S{\'a}nchez, Clara I},
  journal={Medical image analysis},
  volume={42},
  pages={60--88},
  year={2017},
  publisher={Elsevier}
}

@article{rajpurkar2022ai,
  title={AI in health and medicine},
  author={Rajpurkar, Pranav and Chen, Emma and Banerjee, Oishi and Topol, Eric J},
  journal={Nature medicine},
  volume={28},
  number={1},
  pages={31--38},
  year={2022},
  publisher={Nature Publishing Group US New York}
}

@article{dayarathna2024deep,
  title={Deep learning based synthesis of MRI, CT and PET: Review and analysis},
  author={Dayarathna, Sanuwani and Islam, Kh Tohidul and Uribe, Sergio and Yang, Guang and Hayat, Munawar and Chen, Zhaolin},
  journal={Medical image analysis},
  volume={92},
  pages={103046},
  year={2024},
  publisher={Elsevier}
}

@inproceedings{liu2020towards,
  title={Towards faster and stabilized gan training for high-fidelity few-shot image synthesis},
  author={Liu, Bingchen and Zhu, Yizhe and Song, Kunpeng and Elgammal, Ahmed},
  booktitle={International conference on learning representations},
  year={2020}
}

@article{gao2023synthetic,
  title={Synthetic data accelerates the development of generalizable learning-based algorithms for X-ray image analysis},
  author={Gao, Cong and Killeen, Benjamin D and Hu, Yicheng and Grupp, Robert B and Taylor, Russell H and Armand, Mehran and Unberath, Mathias},
  journal={Nature Machine Intelligence},
  volume={5},
  number={3},
  pages={294--308},
  year={2023},
  publisher={Nature Publishing Group UK London}
}

@inproceedings{chen2024towards,
  title={Towards generalizable tumor synthesis},
  author={Chen, Qi and Chen, Xiaoxi and Song, Haorui and Xiong, Zhiwei and Yuille, Alan and Wei, Chen and Zhou, Zongwei},
  booktitle={Proceedings of the IEEE/CVF Conference on Computer Vision and Pattern Recognition},
  pages={11147--11158},
  year={2024}
}

@misc{shang2024synfundus1mhighqualitymillionscalesynthetic,
      title={SynFundus-1M: A High-quality Million-scale Synthetic fundus images Dataset with Fifteen Types of Annotation}, 
      author={Fangxin Shang and Jie Fu and Yehui Yang and Haifeng Huang and Junwei Liu and Lei Ma},
      year={2024},
      eprint={2312.00377},
      archivePrefix={arXiv},
      primaryClass={cs.CV},
      url={https://arxiv.org/abs/2312.00377}, 
}

@article{shumailov2024ai,
  title={AI models collapse when trained on recursively generated data},
  author={Shumailov, Ilia and Shumaylov, Zakhar and Zhao, Yiren and Papernot, Nicolas and Anderson, Ross and Gal, Yarin},
  journal={Nature},
  volume={631},
  number={8022},
  pages={755--759},
  year={2024},
  publisher={Nature Publishing Group UK London}
}

@inproceedings{singh2024synthetic,
  title={Is Synthetic Data all We Need? Benchmarking the Robustness of Models Trained with Synthetic Images},
  author={Singh, Krishnakant and Navaratnam, Thanush and Holmer, Jannik and Schaub-Meyer, Simone and Roth, Stefan},
  booktitle={Proceedings of the IEEE/CVF Conference on Computer Vision and Pattern Recognition},
  pages={2505--2515},
  year={2024}
}

@article{wang2024self,
  title={Self-improving generative foundation model for synthetic medical image generation and clinical applications},
  author={Wang, Jinzhuo and Wang, Kai and Yu, Yunfang and Lu, Yuxing and Xiao, Wenchao and Sun, Zhuo and Liu, Fei and Zou, Zixing and Gao, Yuanxu and Yang, Lei and others},
  journal={Nature Medicine},
  pages={1--9},
  year={2024},
  publisher={Nature Publishing Group}
}

@article{guillaudeux2023patient,
  title={Patient-centric synthetic data generation, no reason to risk re-identification in biomedical data analysis},
  author={Guillaudeux, Morgan and Rousseau, Olivia and Petot, Julien and Bennis, Zineb and Dein, Charles-Axel and Goronflot, Thomas and Vince, Nicolas and Limou, Sophie and Karakachoff, Matilde and Wargny, Matthieu and others},
  journal={NPJ Digital Medicine},
  volume={6},
  number={1},
  pages={37},
  year={2023},
  publisher={Nature Publishing Group UK London}
}

@article{ktena2024generative,
  title={Generative models improve fairness of medical classifiers under distribution shifts},
  author={Ktena, Ira and Wiles, Olivia and Albuquerque, Isabela and Rebuffi, Sylvestre-Alvise and Tanno, Ryutaro and Roy, Abhijit Guha and Azizi, Shekoofeh and Belgrave, Danielle and Kohli, Pushmeet and Cemgil, Taylan and others},
  journal={Nature Medicine},
  pages={1--8},
  year={2024},
  publisher={Nature Publishing Group US New York}
}

@article{xu2019frequency,
  title={Frequency principle: Fourier analysis sheds light on deep neural networks},
  author={Xu, Zhi-Qin John and Zhang, Yaoyu and Luo, Tao and Xiao, Yanyang and Ma, Zheng},
  journal={arXiv preprint arXiv:1901.06523},
  year={2019}
}

@inproceedings{wang2020high,
  title={High-frequency component helps explain the generalization of convolutional neural networks},
  author={Wang, Haohan and Wu, Xindi and Huang, Zeyi and Xing, Eric P},
  booktitle={Proceedings of the IEEE/CVF conference on computer vision and pattern recognition},
  pages={8684--8694},
  year={2020}
}

@inproceedings{rahaman2019spectral,
  title={On the spectral bias of neural networks},
  author={Rahaman, Nasim and Baratin, Aristide and Arpit, Devansh and Draxler, Felix and Lin, Min and Hamprecht, Fred and Bengio, Yoshua and Courville, Aaron},
  booktitle={International conference on machine learning},
  pages={5301--5310},
  year={2019},
  organization={PMLR}
}

@inproceedings{durall2020watch,
  title={Watch your up-convolution: Cnn based generative deep neural networks are failing to reproduce spectral distributions},
  author={Durall, Ricard and Keuper, Margret and Keuper, Janis},
  booktitle={Proceedings of the IEEE/CVF conference on computer vision and pattern recognition},
  pages={7890--7899},
  year={2020}
}

@article{dzanic2020fourier,
  title={Fourier spectrum discrepancies in deep network generated images},
  author={Dzanic, Tarik and Shah, Karan and Witherden, Freddie},
  journal={Advances in neural information processing systems},
  volume={33},
  pages={3022--3032},
  year={2020}
}

@INPROCEEDINGS {9577744,
author = { Liu, Honggu and Li, Xiaodan and Zhou, Wenbo and Chen, Yuefeng and He, Yuan and Xue, Hui and Zhang, Weiming and Yu, Nenghai },
booktitle = { 2021 IEEE/CVF Conference on Computer Vision and Pattern Recognition (CVPR) },
title = {{ Spatial-Phase Shallow Learning: Rethinking Face Forgery Detection in Frequency Domain }},
year = {2021},
volume = {},
ISSN = {},
pages = {772-781},
doi = {10.1109/CVPR46437.2021.00083},
url = {https://doi.ieeecomputersociety.org/10.1109/CVPR46437.2021.00083},
publisher = {IEEE Computer Society},
address = {Los Alamitos, CA, USA},
month =Jun}

@inproceedings{corvi2023intriguing,
  title={Intriguing properties of synthetic images: from generative adversarial networks to diffusion models},
  author={Corvi, Riccardo and Cozzolino, Davide and Poggi, Giovanni and Nagano, Koki and Verdoliva, Luisa},
  booktitle={Proceedings of the IEEE/CVF Conference on Computer Vision and Pattern Recognition},
  pages={973--982},
  year={2023}
}

@inproceedings{frank2020leveraging,
  title={Leveraging frequency analysis for deep fake image recognition},
  author={Frank, Joel and Eisenhofer, Thorsten and Sch{\"o}nherr, Lea and Fischer, Asja and Kolossa, Dorothea and Holz, Thorsten},
  booktitle={International conference on machine learning},
  pages={3247--3258},
  year={2020},
  organization={PMLR}
}

@inproceedings{jiang2021focal,
  title={Focal frequency loss for image reconstruction and synthesis},
  author={Jiang, Liming and Dai, Bo and Wu, Wayne and Loy, Chen Change},
  booktitle={Proceedings of the IEEE/CVF international conference on computer vision},
  pages={13919--13929},
  year={2021}
}

@inproceedings{zamir2022restormer,
  title={Restormer: Efficient transformer for high-resolution image restoration},
  author={Zamir, Syed Waqas and Arora, Aditya and Khan, Salman and Hayat, Munawar and Khan, Fahad Shahbaz and Yang, Ming-Hsuan},
  booktitle={Proceedings of the IEEE/CVF conference on computer vision and pattern recognition},
  pages={5728--5739},
  year={2022}
}

@article{hochreiter1997long,
  title={Long Short-term Memory},
  author={Hochreiter, S},
  journal={Neural Computation MIT-Press},
  year={1997}
}

@misc{ilanchezian2023generatingrealisticcounterfactualsretinal,
      title={Generating Realistic Counterfactuals for Retinal Fundus and OCT Images using Diffusion Models}, 
      author={Indu Ilanchezian and Valentyn Boreiko and Laura Kühlewein and Ziwei Huang and Murat Seçkin Ayhan and Matthias Hein and Lisa Koch and Philipp Berens},
      year={2023},
      eprint={2311.11629},
      archivePrefix={arXiv},
      primaryClass={cs.CV},
      url={https://arxiv.org/abs/2311.11629}, 
}

@inproceedings{Karras2021,
  author = {Tero Karras and Miika Aittala and Samuli Laine and Erik H\"ark\"onen and Janne Hellsten and Jaakko Lehtinen and Timo Aila},
  title = {Alias-Free Generative Adversarial Networks},
  booktitle = {Proc. NeurIPS},
  year = {2021}
}

@inproceedings{cubuk2019autoaugment,
  title={Autoaugment: Learning augmentation strategies from data},
  author={Cubuk, Ekin D and Zoph, Barret and Mane, Dandelion and Vasudevan, Vijay and Le, Quoc V},
  booktitle={Proceedings of the IEEE/CVF conference on computer vision and pattern recognition},
  pages={113--123},
  year={2019}
}

@article{kang2023studiogan,
  title={StudioGAN: a taxonomy and benchmark of GANs for image synthesis},
  author={Kang, Minguk and Shin, Joonghyuk and Park, Jaesik},
  journal={IEEE Transactions on Pattern Analysis and Machine Intelligence},
  volume={45},
  number={12},
  pages={15725--15742},
  year={2023},
  publisher={IEEE}
}

@inproceedings{karras2020analyzing,
  title={Analyzing and improving the image quality of stylegan},
  author={Karras, Tero and Laine, Samuli and Aittala, Miika and Hellsten, Janne and Lehtinen, Jaakko and Aila, Timo},
  booktitle={Proceedings of the IEEE/CVF conference on computer vision and pattern recognition},
  pages={8110--8119},
  year={2020}
}

@InProceedings{Psaroudakis_2022_CVPR,
    author    = {Psaroudakis, Andreas and Kollias, Dimitrios},
    title     = {MixAugment \& Mixup: Augmentation Methods for Facial Expression Recognition},
    booktitle = {Proceedings of the IEEE/CVF Conference on Computer Vision and Pattern Recognition (CVPR) Workshops},
    month     = {June},
    year      = {2022},
    pages     = {2367-2375}
}

@INPROCEEDINGS{10655510,
  author={Vaish, Puru and Wang, Shunxin and Strisciuglio, Nicola},
  booktitle={2024 IEEE/CVF Conference on Computer Vision and Pattern Recognition (CVPR)}, 
  title={Fourier-Basis Functions to Bridge Augmentation Gap: Rethinking Frequency Augmentation in Image Classification}, 
  year={2024},
  volume={},
  number={},
  pages={17763-17772},
  doi={10.1109/CVPR52733.2024.01682}}

@INPROCEEDINGS{10657264,
  author={Wang, Yinong Oliver and Chung, Younjoon and Wu, Chen Henry and De la Torre, Fernando},
  booktitle={2024 IEEE/CVF Conference on Computer Vision and Pattern Recognition (CVPR)}, 
  title={Domain Gap Embeddings for Generative Dataset Augmentation}, 
  year={2024},
  volume={},
  number={},
  pages={28684-28694},
  doi={10.1109/CVPR52733.2024.02710}}

@misc{sartaj_bhuvaji_ankita_kadam_prajakta_bhumkar_sameer_dedge_swati_kanchan_2020,
	title={Brain Tumor Classification (MRI)},
	url={https://www.kaggle.com/dsv/1183165},
	DOI={10.34740/KAGGLE/DSV/1183165},
	publisher={Kaggle},
	author={Sartaj Bhuvaji and Ankita Kadam and Prajakta Bhumkar and Sameer Dedge and Swati Kanchan},
	year={2020}
}

@misc{diabetic-retinopathy-detection,
    author = {Emma Dugas and Jared and Jorge and Will Cukierski},
    title = {Diabetic Retinopathy Detection},
    year = {2015},
    howpublished = {\url{https://kaggle.com/competitions/diabetic-retinopathy-detection}},
    note = {Kaggle}
}

@article{johnsonMIMICCXRDeidentifiedPublicly2019,
  title = {{{MIMIC-CXR}}, a de-Identified Publicly Available Database of Chest Radiographs with Free-Text Reports},
  author = {Johnson, Alistair E. W. and Pollard, Tom J. and Berkowitz, Seth J. and Greenbaum, Nathaniel R. and Lungren, Matthew P. and Deng, Chih-ying and Mark, Roger G. and Horng, Steven},
  year = {2019},
  month = dec,
  journal = {Scientific Data},
  volume = {6},
  number = {1},
  pages = {317},
  issn = {2052-4463},
  doi = {10.1038/s41597-019-0322-0},
  langid = {english}
}

@article{wang2024economic,
  title={Economic evaluation for medical artificial intelligence: accuracy vs. cost-effectiveness in a diabetic retinopathy screening case},
  author={Wang, Yueye and Liu, Chi and Hu, Wenyi and Luo, Lixia and Shi, Danli and Zhang, Jian and Yin, Qiuxia and Zhang, Lei and Han, Xiaotong and He, Mingguang},
  journal={NPJ Digital Medicine},
  volume={7},
  number={1},
  pages={43},
  year={2024},
  publisher={Nature Publishing Group UK London}
}

@article{wang2024towards,
  title={Towards regulatory generative AI in ophthalmology healthcare: a security and privacy perspective},
  author={Wang, Yueye and Liu, Chi and Zhou, Keyao and Zhu, Tianqing and Han, Xiaotong},
  journal={British Journal of Ophthalmology},
  volume={108},
  number={10},
  pages={1349--1353},
  year={2024},
  publisher={BMJ Publishing Group Ltd}
}

@article{li2018automated,
  title={An automated grading system for detection of vision-threatening referable diabetic retinopathy on the basis of color fundus photographs},
  author={Li, Zhixi and Keel, Stuart and Liu, Chi and He, Yifan and Meng, Wei and Scheetz, Jane and Lee, Pei Ying and Shaw, Jonathan and Ting, Daniel and Wong, Tien Yin and others},
  journal={Diabetes care},
  volume={41},
  number={12},
  pages={2509--2516},
  year={2018},
  publisher={American Diabetes Association}
}

@article{he2020deployment,
  title={Deployment of artificial intelligence in real-world practice: opportunity and challenge},
  author={He, Mingguang and Li, Zhixi and Liu, Chi and Shi, Danli and Tan, Zachary},
  journal={Asia-Pacific Journal of Ophthalmology},
  volume={9},
  number={4},
  pages={299--307},
  year={2020},
  publisher={Elsevier}
}
